\documentclass[a4paper,fleqn]{cas-dc}

\usepackage[authoryear,longnamesfirst]{natbib}
\usepackage{rotating}
\usepackage{colortbl}
\usepackage{xcolor}
\usepackage{pgfplots}
\usepackage{collcell}
\usepackage{amsmath}
\newtheorem{lemma}{Lemma}

\def\tsc#1{\csdef{#1}{\textsc{\lowercase{#1}}\xspace}}
\tsc{WGM}
\tsc{QE}

\begin{document}
\let\WriteBookmarks\relax
\def\floatpagepagefraction{1}
\def\textpagefraction{.001}

\shorttitle{CANet}    

\shortauthors{Sonmezer \& Ertekin}  

\title [mode = title]{CANet: ChronoAdaptive Network for Enhanced Long-Term Time Series Forecasting under Non-Stationarity}



%

\author[1]{Mert Sonmezer}[orcid=0009-0004-1016-2449]

\cormark[1]


\ead{mert.sonmezer@metu.edu.tr}


\credit{Writing - original draft, Visualization, Validation, Methodology, Software, Conceptualization, Formal analysis, Data curation, Investigation}

\affiliation[1]{organization={Department of Computer Engineering},
            addressline={Middle East Technical University}, 
            city={Ankara},
            postcode={06800},
            country={Turkey}}

\author[1, 2]{Seyda Ertekin}[orcid=0000-0002-6132-6739]


\ead{sertekin@metu.edu.tr}


\credit{Writing - review, Methodology, Conceptualization, Supervision, Project administration, Resources}

\affiliation[2]{organization={METU-BILTIR CAD/CAM \& Robotics Research Center},
            addressline={Middle East Technical University}, 
            city={Ankara},
            postcode={06800},
            country={Turkey}}

\cortext[1]{Corresponding author at: Department of Computer Engineering, Middle East Technical University, Ankara, 06800, Turkey}



\begin{abstract}
Long-term time series forecasting plays a pivotal role in various real-world applications. Despite recent advancements and the success of different architectures, forecasting is often challenging due to non-stationary nature of the real-world data, which frequently exhibit distribution shifts and temporal changes in statistical properties like mean and variance over time. Previous studies suggest that this inherent variability complicates forecasting, limiting the performance of many models by leading to loss of non-stationarity and resulting in over-stationarization \citep{Liu2022}. In order to address this challenge, we introduce a novel architecture, \textbf{C}horono\textbf{A}daptive \textbf{Net}work (CANet), inspired by style-transfer techniques. The core of CANet is the Non-stationary Adaptive Normalization module, seamlessly integrating the Style Blending Gate and Adaptive Instance Normalization (AdaIN) \citep{Huang2017}. The Style Blending Gate preserves and reintegrates non-stationary characteristics, such as mean and standard deviation, by blending internal and external statistics, preventing over-stationarization while maintaining essential temporal dependencies. Coupled with AdaIN, which dynamically adapts the model to statistical changes, this approach enhances predictive accuracy under non-stationary conditions. CANet also employs multi-resolution patching to handle short-term fluctuations and long-term trends, along with Fourier analysis-based adaptive thresholding to reduce noise. A Stacked Kronecker Product Layer further optimizes the model's efficiency while maintaining high performance. Extensive experiments on real-world datasets validate CANet's superiority over state-of-the-art methods, achieving a 42\% reduction in MSE and a 22\% reduction in MAE. The source code is publicly available.\footnotemark.
\end{abstract}


\nocite{*}


\maketitle

\section{Introduction}
\label{sec:intro}
\footnotetext[1]{https://github.com/mertsonmezer/CANet}

Time series forecasting is extensively applied in practical domains such as finance, healthcare, electricity, transportation, and weather forecasting. Recently, the Transformer model \citep{Vaswani2017}, initially celebrated for its innovations in natural language processing, has been effectively repurposed for time series analysis. This adaptation leverages its capacity to capture long-range dependencies within time series, demonstrating adeptness in forecasting applications \citep{Wu2021, Zhou2022, liu2024itransformer}. However, despite early achievements in forecasting, Transformers face challenges in varied time series applications, especially with smaller datasets, due to their quadratic complexity and large number of parameters, which may result in overfitting and computational inefficiencies \citep{Wen2023}. Additionally, the attention mechanism of Transformers often struggles with noise and redundancy typical of time series data \citep{Li2022}.

To address these limitations of Transformers, researchers have developed models based on Convolutional Neural Networks (CNNs) \citep{Eldele2024, Donghao2024, Shen2023}. Owing to their lightweight structure, CNN-based models are less prone to overfitting and computational inefficiency compared to Transformers. Although traditional CNNs struggled with capturing long-term dependencies, recent innovations such as the Depthwise Convolution (DWConv) \citep{Donghao2024} and the Adaptive Spectral Block (ASB) \citep{Eldele2024} have shown comparable performance to the attention mechanism, without the associated drawbacks of noise and redundancy. These advancements are proving CNN-based models to be competitive alternatives to Transformer-based models in time series forecasting tasks.

Despite architectural progress, handling the non-stationary nature of real-world time series remains a core challenge. Non-stationary time series are marked by continuously evolving statistical properties and joint distributions, which complicate modeling and reduce generalization capacity \citep{Anderson1976, Hyndman2021, Pan2010, Li2017, Ahuja2021}. To address this, prior work has explored two main directions: normalization-based approaches and invariance-learning approaches. Normalization methods \citep{Ogasawara2010, Kim2022, Liu2023, ye2024, fan2023dishtsgeneralparadigmalleviating} aim to stabilize training by standardizing inputs or activations using either global or sample-specific statistics, acting as preprocessing or model-agnostic layers to reduce distribution shifts. In contrast, invariance-learning methods seek representation-level robustness by learning features that remain stable under certain transformations or regime changes, such as value offsets, trends, or other environment-induced shifts\citep{pmlr-v267-germain25a, liu2024foil}. These approaches follow principles from causality and out-of-distribution (OOD) generalization by explicitly discouraging the model from relying on spurious, time-varying correlations. While both families of methods reduce a model's sensitivity to distribution shifts, they do so via different mechanisms: normalization removes variability through statistical standardization, whereas invariance learning limits reliance on variability by reshaping the learned feature space. This distinction is important yet often overlooked. Moreover, both strategies can inadvertently suppress meaningful temporal signals if not carefully designed, leading to the over-stationarization effect.

\begin{figure*}[]
    \centering
    \includegraphics[width=0.9\linewidth]{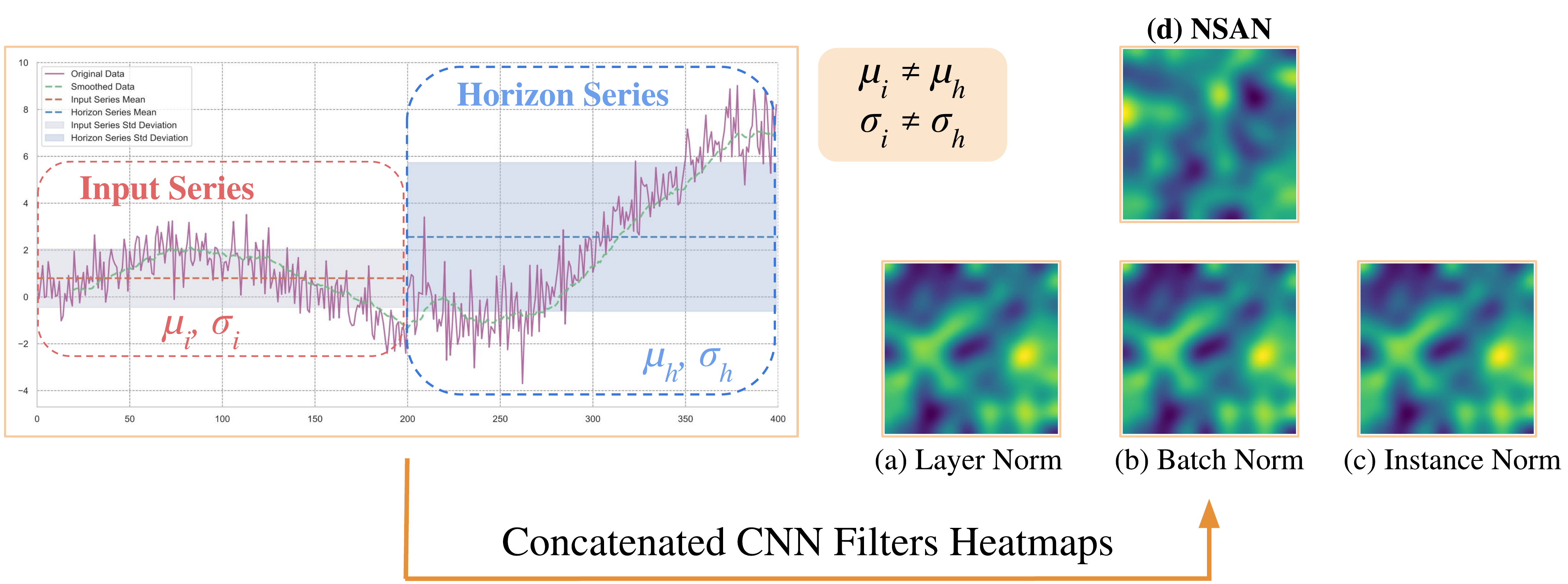}
    \caption[]{Visualization of over-stationarization effects on the internal convolutional filter representations across four normalization methods, using identical model configurations trained on the Exchange dataset. (a)–(c) show that Layer, Batch, and Instance Normalization yield highly similar learned filters, consistent with their standardizing effect (see Lemma \ref{lemma:over-stationarization}). In contrast, (d) our NSAN learns a different filter configuration, reflecting the influence of its blended instance-specific statistics. This distinct internal parameterization aligns with NSAN's consistently superior ablation performance (see Table \ref{tbl:ablation}), supporting its role in mitigating over-stationarization and improving forecasting accuracy.}
    \vspace{-5mm}
    \label{fig:feature-maps}
\end{figure*}

Neverthless, non-stationarity is not just a challenge but also an intrinsic characteristic of real-world time series, potentially serving as a valuable indicator for uncovering essential temporal dependencies vital for forecasting. Recognizing this, recent studies, such as the work by \citet{Liu2022}, points out that traditional preprocessing methods may inadvertently lead to over-stationarization, erasing meaningful variations needed for accurate prediction. These methods typically operate at the high-level level and assume global or sample-invariant statistics, without adjusting to the evolving internal representations across layers. Additionally, our experimental findings suggest that incorporating the non-stationary characteristics of time series data into the model leads to distinct internal filter configurations, as illustrated in Figure \ref{fig:feature-maps}, bringing previously unrecognized aspects of the data to the forefront. This observation raises a critical question: \textit{How can we design a lightweight CNN-based model that avoids the pitfalls of over-stationarization?}

In this paper, we introduce \textbf{C}hrono\textbf{A}daptive \textbf{Net}work (CANet), a novel architecture that enhances CNNs and their efficient variants to adeptly manage the non-stationary nature of real-world time series data. The model initially employs multi-dimensional patches \citep{Nie2023, Zhang2024} as inputs, considered to be fundamental units of time series data due to their richer informational content than a single point \citep{Cheng2023, He2023}. It utilizes the Adaptive Spectral Block (ASB) and Interactive Convolutional Block (ICB) \citep{Eldele2024} to mitigate noise through Fourier-based multiplications using global and local filters, capture both long and short-term dependencies, and enable interpretation of complex temporal patterns. In order to combat the challenge of over-stationarization and enhance the predictability of non-stationary time series, CANet features Non-stationary Adaptive Normalization (NSAN), which operates within the model itself, integrating the Style Blending Gate with Adaptive Instance Normalization (AdaIN) \citep{Huang2017}. This normalization method dynamically adapts to statistical changes in the data, using the input's mean and variance as style inputs to preserve intrinsic data characteristics and boost predictive accuracy amid temporal variations. Furthermore, the incorporation of a Stacked Kronecker Product Layer \citep{Panahi2021} ensures that CANet remains lightweight and easily deployable, ideal for practical applications.

In summary, our primary contributions are as follows:

\begin{itemize}
    \item We present CANet, a versatile and lightweight model architecture specifically designed for a wide range of time series forecasting tasks.
    \item We introduce the integration of the Style Blending Gate with AdaIN within the NSAN module, providing a computationally efficient approach to maintain the inherent non-stationarity of the data, thereby addressing the issue of over-stationarization.
    \item CANet demonstrates superior performance against various leading time series forecasting models in multivariate long-term forecasting tasks.
\end{itemize}

The paper is organized as follows: Section \ref{sec:related-works} reviews related studies, providing a theoretical basis and context for this work. Section \ref{sec:proposed-method} details the development and unique elements of CANet, highlighting its design principles and intended benefits. Section \ref{sec:experiments} outlines the experimental setup, including data sources and the testing environment, and presents a comparative analysis of CANet’s performance against baseline models along with ablation studies to evaluate model design choices. Lastly, Section \ref{sec:conclusion} summarizes the primary contributions and offers insights for future research directions.

\section{Related work}
\label{sec:related-works}

\subsection{Deep models for time series forecasting}
\label{subsec:deep-models}

\subsubsection{RNN and MLP-based models}
\label{subsubsec:mlp-rnn-based-models}
Recent advancements in time series forecasting have been propelled by the development of both meticulously designed architectures and sophisticated algorithms. Initially, techniques such as RNN-based methods \citep{Hochreiter1997, Wen2017, Yu2017, Maddix2018, Rangapuram2018, Salinas2020} were widely adopted, leveraging their recurrent design and memory mechanisms to uncover implicit temporal transitions. Despite their promising beginnings, these methods encountered difficulties with capturing long-range dependencies and cumulative error propagation. On the other hand, models based on multilayer perceptrons (MLPs) \citep{Oreshkin2020, Zeng2023, zhang2022morefastmultivariatetime}, which learn weights independently at each time step, have shown remarkable performance, demonstrating their effectiveness in capturing simpler temporal patterns. Nevertheless, their practical use remains limited for non-stationary time series, whose time-dependent characteristics challenge the models' capacity and efficiency.

\subsubsection{Transformer-based models}
\label{subsubsec:transformer-based-models}

Transformer-based models became popular in time series forecasting due to their proficiency in managing long-term dependencies via the attention mechanism. Models like Informer \citep{Zhou2021} and FEDformer \citep{Zhou2022} use low-rank matrices in self-attention mechanisms to handle long sequences better. Reformer \citep{Kitaev2020} utilizes locally sensitive hashing to approximate attention, while Autoformer \citep{Wu2021} employs block decomposition and autocorrelation for capturing time series characteristics. Similarly, iTransformer \citep{liu2024itransformer} integrates attention with feed-forward networks to model multivariate correlations and nonlinear patterns. MTST \citep{Zhang2024} leverage multi-resolution patches to capture both long-range and local dependencies effectively.  Furthermore, the Non-stationary Transformer \citep{Liu2022} employs a de-stationary attention to tackle over-stationarization in time series. Nevertheless, recent analyses \citep{Kim2022, Kuznetsov2020} point out persistent challenges with Transformers, including their difficulty in generalizing across dynamic temporal distributions, high computational demands, and a tendency to misrepresent temporal data \citep{Zeng2023}.

\subsubsection{CNN-based models} 
\label{subsubsec:cnn-based-models}

CNNs have gained popularity in the field, especially with the models like TCN and its variants \citep{Bai2018, Sen2019, Franceschi2019} that use causal convolution to capture temporal causality. Nevertheless, these models often struggle to capture long-range dependencies, prompting further innovations. For example, MICN \citep{Wang2023} enhances traditional CNN-based models by introducing a multi-scale convolution structure that integrates local features with global correlations.  SCINet \citep{Liu2022Scinet} abandons causal convolution in favor of a recursive downsample-convolve-interact architecture designed to handle complex temporal dynamics. Additionally, T-WaveNet \citep{Liu2022T-wavenet} utilizes frequency spectrum energy analysis for more effective signal decomposition, and WFTNet \citep{Liu2024Wftnet} combines Fourier and wavelet transforms to provide comprehensive temporal-frequency analysis, marking progress toward more accurate and efficient time series forecasting models. Similarly, TSLANet \citep{Eldele2024} introduces an Adaptive Spectral Block using Fourier analysis to enhance feature representation and capture both long-term and short-term dependencies, while mitigating noise through adaptive thresholding, establishing itself as a robust and efficient CNN-based alternative for diverse time series tasks.

\subsection{Handling Non-Stationarity in Time Series Forecasting}
\label{subsec:non-stationary-time-series}

Non-stationarity presents a fundamental challenge for real-world forecasting tasks. It causes distribution shifts between training and test data, weakens the generalization of the model, and complicates long-term prediction. Various strategies have been proposed in the literature to mitigate these effects, ranging from statistical preprocessing techniques to neural models designed to explicitly handle distributional changes.

Historically, ARIMA \citep{Geurts1977, Box1968}, a classical statistical method, attempted to stationarize time series through differencing. In deep learning, given the challenges posed by changing distributions due to non-stationarity, normalizing the input data has become the most commonly adopted method to mitigate these effects. For example, Adaptive Normalization \citep{Ogasawara2010} uses z-score normalization on each series fragment using global statistics from a sampled set. DAIN \citep{Passalis2020} introduces a non-linear network that adaptively normalizes each input to align with the output's statistics, while ST-Norm  \citep{Deng2021} proposed separate modules to normalize spatial and temporal axes in multivariate series. RevIN \citep{Kim2022} advanced this idea by offering reversible normalization that operates both at the input and output stages, effectively isolating trend and scale while preserving reconstruction fidelity.

However, these normalization schemes are generally designed as model-agnostic modules: they operate at the boundaries of the forecasting backbone (i.e., on the input or output sequences) rather than within its feature-extraction layers. In this line, SAN \citep{Liu2023} introduces temporal slicing, normalizing sub-series instead of whole sequences and learning slice-specific statistics via a statistics-prediction network. FAN \citep{ye2024} extends this idea to the frequency domain, identifying non-stationary frequency components through Fourier analysis and adapting them with auxiliary prediction heads. Dish-TS \citep{fan2023dishtsgeneralparadigmalleviating} similarly learns level and scale coefficients for both lookback and horizon windows using auxiliary neural networks, and DAIN \citep{Passalis2020} employs a deep input-normalization layer that adaptively shifts and scales each series before it enters the forecasting model. Thus, while approaches such as DAIN, SAN, FAN, and Dish-TS implement learnable, sample-adaptive normalization trained jointly with the model, their effects are primarily pre- or post-processing: they indirectly shape the internal representations seen by the backbone by altering its input. In contrast, our NSAN module is implemented as an internal normalization layer within CANet's convolutional blocks and directly modulates intermediate feature maps via a blend of global instance-wise statistics and local contextual style. This layer-wise integration injects non-stationary statistics at multiple depths of the network and is explicitly designed to mitigate feature-space over-stationarization from within the backbone itself, an effect we formalize in Section~\ref{subsubsec:style-blending-gate}.

Beyond normalization, several works attempt to model non-stationary patterns within the architecture. The Non-stationary Transformer \citep{Liu2022} introduced the De-stationary Attention mechanism to adaptively capture temporal shifts. Koopa \citep{liu2023koopa} applied Koopman operator theory to decompose dynamics into invariant and evolving components. Additionally, a growing line of research has explored invariance-learning approaches that explicitly enforce robustness to transformations such as offsets, trends, or environment-induced shifts. InvConvNet \citep{pmlr-v267-germain25a} learns representations invariant to offsets and linear trends via CNN-based adaptations, while FOIL \citep{liu2024foil} promotes environment-invariant representations through latent domain inference and a surrogate multi-head loss. LT-normalized distance \citep{germain-ltnormalized} extends classical z-normalization by modeling invariance to linear trends. These methods follow the broader principle of shaping representations to remain stable under specific deformations, connecting non-stationarity modeling to the literature on decomposition, causality, and OOD generalization. However, many of these approaches rely on complex auxiliary objectives or computationally intensive backbones, which can lead to higher training overhead and training instability.

To bridge these gaps, CANet introduces a fully convolutional architecture with an integrated NSAN module. In contrast to model-agnostic normalization techniques, NSAN is embedded directly into the model and modulates intermediate feature activations based on both global instance-wise statistics and local contextual style, enabled via a Style Blending Gate and AdaIN mechanism. This design allows CANet to dynamically respond to distributional shifts across layers while preserving essential temporal variations. Moreover, CANet remains lightweight and efficient, requiring no transformers, attention heads, or environment annotations, and achieves strong performance particularly on highly non-stationary benchmarks.

\section{Proposed method: CANet}
\label{sec:proposed-method}

\subsection{Problem statement}
\label{subsec:problem-statement}

This paper concentrates on forecasting multivariate time series. Let $X$ represent such a series. The task of multivariate time series forecasting at time step $t$ is to predict the target time series $X_{t+1:t+O} = \{X_{t+1}, ..., X_{t+O}\}$ in the future $O$ time steps based on the past time series $X_{t-L+1:t} = \{X_{t-L+1:t}, ..., X_t\}$ of length $L$, where $X_t \in R^n$ is the value of the time series at time step $t$, $n$ is the number of variables, and $O$ is the prediction horizon. The forecasting function $f$ can be formalized as follows:

\begin{equation}
    \hat{X}_{t+1:t+O} = f(X_{t-L+1:t})
\end{equation}

\noindent where  $\hat{X}_{t+1:t+O}$ is the predicted time series output from the model,  and $f$ is the nonlinear function to be learned. In this work, we focus on the long-term forecasting of multivariate time series, where $n > 1$, and adopt a multi-step approach to predict the full forecast horizon in a single forward pass.

\begin{figure*}[]
  \centering
    \includegraphics[width=0.9\linewidth]{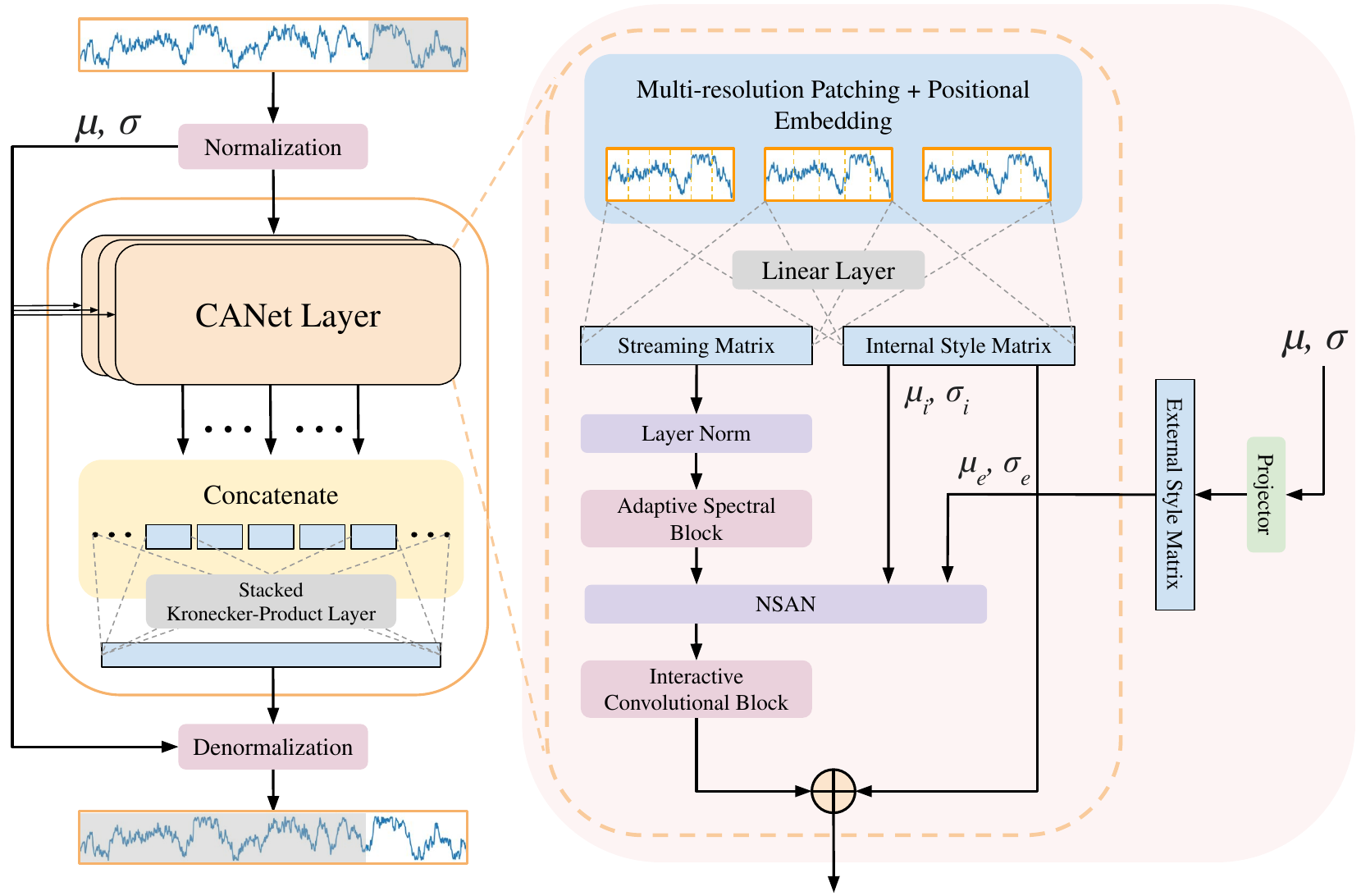}
    \caption[]{The overall architecture of CANet. Instance Normalization encapsulates the base model, normalizing each input series and then denormalizing it right before output generation \citep{Kim2022}. The NSAN module is employed to reintroduce diverse patterns associated with the non-stationary characteristics of time series data.}
    \vspace{-5mm}
    \label{fig:model-architecture}
\end{figure*}

\subsection{Overall framework}
\label{subsec:overall-framework}

Non-stationarity is commonly observed in real-world time series data as previously noted. Traditional methods that simply normalize and denormalize data often fail to take into account the dynamic, non-stationary characteristics of time series, leading to what is known as over-stationarization, as illustrated in Figure \ref{fig:feature-maps}. In order to effectively capture the unique features arising from non-stationary traits and tackle over-stationarization, a new framework, ChronoAdaptive Network (CANet), has been developed, of which architecture is illustrated in Figure \ref{fig:model-architecture}.

This framework takes inspiration from TSLANet \citep{Eldele2024}, a leading model in the field of multivariate long-term time series forecasting known for its accuracy and ease of deployment. Recognizing TSLANet’s capacity for noise reduction and its lightweight structure, improvements were made by incorporating multi-resolution patching and the Non-stationary Adaptive Normalization (NSAN) module, of which internal structure is shown in Figure \ref{fig:nsan}. This module features a Style Blending Gate and Adaptive Instance Normalization (AdaIN), which reintegrates the non-stationary information from the raw series into the model.

CANet operates through several key steps. Initially, it normalizes the incoming time series data, extracting the mean and standard deviation which are then fed back into the model at subsequent stages. It then proceeds to generate two matrices through multi-resolution patching: one called the internal style matrix, and the other serving as the primary matrix for prediction. The internal style matrix in CANet plays a critical role by adaptively adjusting the model’s processing to accommodate distribution shifts within the time series data, which is an analogous approach to DAIN's adaptive mechanism \citep{Passalis2020}. By providing this matrix to AdaIN, CANet dynamically modulates its internal representation to better fit non-stationary patterns by recalibrating to changes in the underlying data distribution in real time. This enables the model to handle both stationary and non-stationary segments more robustly, enhancing its adaptability to diverse trends and patterns. The Non-stationary Adaptive Normalization module utilizes both the internal style matrix and an external style matrix—derived from non-stationary aspects following a projection layer—to forecast the time series data.

The integration of these enhancements leads to the development of CANet, a model that aims to simplify and improve the efficiency of deploying multivariate long-term time series forecasting solutions.

\subsection{Non-stationary Adaptive Normalization}
\label{subsec:nsan}

Although the statistics of each time series are restored in the corresponding predictions, merely applying denormalization fails to fully recover the inherent non-stationarity of the original series, as noted by \citet{Liu2022}. This leads to a significant issue: different time series, such as $x^{(1)}$ and $x^{(2)}$, might be transformed into the same stationarized input $x'$ following normalization. Consequently, identical inputs produce indistinguishable feature maps in the CNN modules, a phenomenon referred to as over-stationarization.

\begin{figure*}[]
  \centering
    \includegraphics[width=0.8\linewidth]{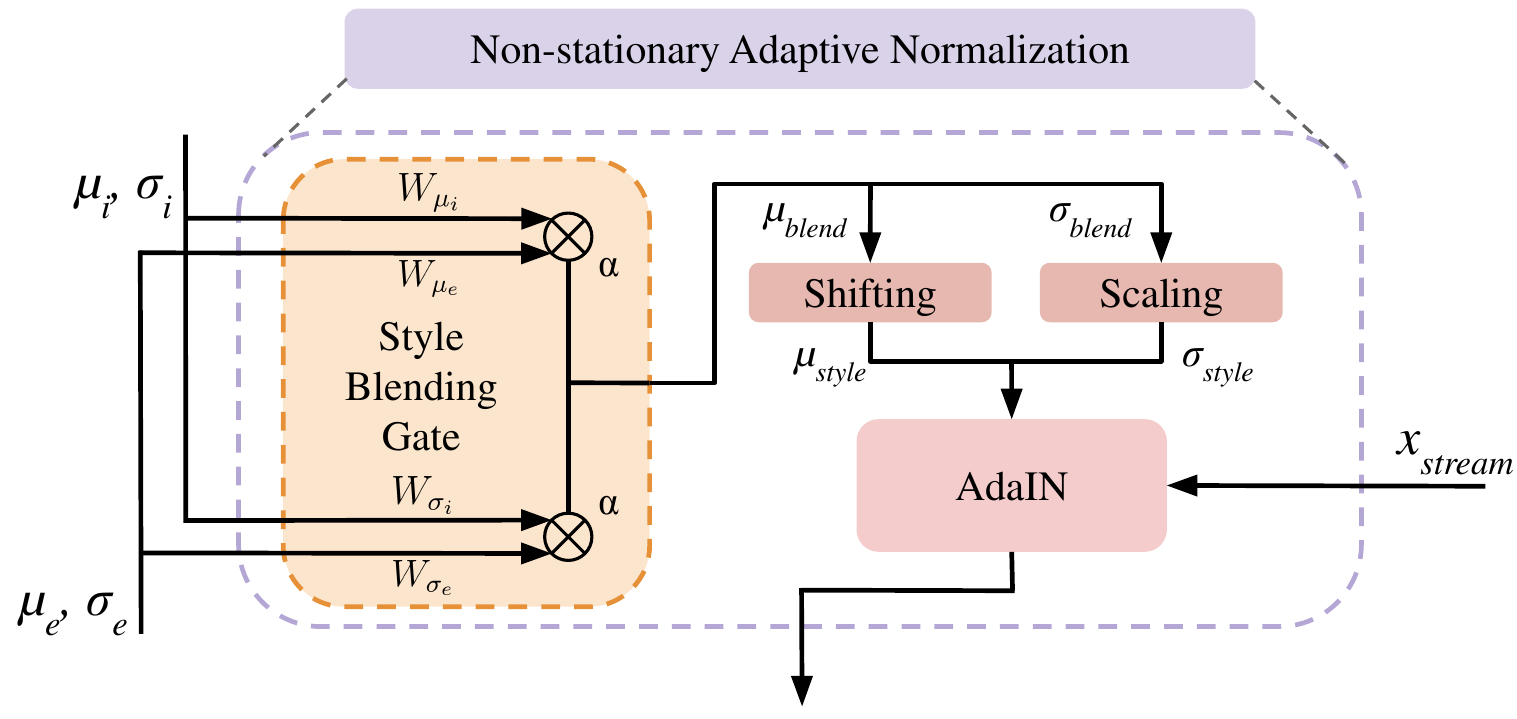}
    \caption[]{The illustration of the Non-stationary Adaptive Normalization (NSAN) module, which integrates the Style Blending Gate with AdaIN to restore the non-stationary characteristics inherent in the original input time series. This module takes the mean and standard deviation as style properties and applies transformations through shifting and scaling operations, enabling the streaming matrix to adjust to changes in data distribution.}
    \vspace{-5mm}
    \label{fig:nsan}
\end{figure*}

This process ultimately results in the feature maps that fail to capture the unique, non-stationary characteristics of each input. In order to remedy this, we have developed the Non-stationary Adaptive Normalization (NSAN) module depicted in Figure \ref{fig:nsan}, which ensures that the model integrates the non-stationary attributes of the original data into its predictions, thereby preventing the production of outputs that overlook these essential non-stationary attributes. NSAN represents the first adaptation of AdaIN to time series forecasting, repurposed here to dynamically modulate internal feature representations based on blended input- and feature-level statistics.

\subsubsection{Style blending gate}
\label{subsubsec:style-blending-gate}

The Style Blending Gate module within the NSAN module is critical for preserving and reintegrating the non-stationary characteristics of the original time series data, such as mean and standard deviation. This module addresses the issue of over-stationarization, where vital non-stationary information is lost, preventing the base model from effectively capturing significant temporal dependencies.

\vspace{\baselineskip}
\noindent\textbf{Mathematical preliminaries.} Let \(N(x)=\frac{x-\mu(x)}{\sigma(x)}\) denote the per-series $z$-score normalizer. We say over-stationarization occurs if

\begin{lemma}
\label{lemma:over-stationarization}
    $\exists\,x^{(1)} \neq x^{(2)} \;\text{s.t.}\; N\!\bigl(x^{(1)}\bigr) = N\!\bigl(x^{(2)}\bigr)$
\end{lemma}

This collapse arises whenever two sequences share identical first-order statistics (mean and variance) yet differ in higher-order moments. When this happens, downstream CNN layers receive indistinguishable feature maps, obscuring meaningful temporal cues.

Formally, the Style Blending Gate processes two sets of statistical properties derived from internal and external style matrices: $\mu_i$, $\sigma_i$ (the mean and standard deviation of the internal style matrix) and $\mu_e$, $\sigma_e$ (the mean and standard deviation of the external style matrix). It applies learnable weight matrices to these properties to modulate their influence before blending:

\begin{equation}
    \begin{split}
        \mu'_i = W_{\mu_i} \cdot \mu_i \qquad\qquad \sigma'_i = W_{\sigma_i} \cdot \sigma_i \\
        \mu'_e = W_{\mu_e} \cdot \mu_e \qquad\qquad \sigma'_e = W_{\sigma_e} \cdot \sigma_e
    \end{split}
\end{equation}

where $W_{\mu_i}$, $W_{\sigma_i}$, $W_{\mu_e}$, and $W_{\sigma_e}$ are learnable parameter matrices that fine-tune the mean and variance from both style matrices to better represent the non-stationary features of the original data.

Once adjusted, these means and variances are blended using a hyperparameter $\alpha$ called blend ratio, which determines the proportion of each style's contribution::

\begin{equation}
    \begin{split}
        \mu_{\textit{blend}} = \alpha \cdot \mu'_i + (1 - \alpha) \cdot \mu'_e \\
        \sigma_{\textit{blend}} = \alpha \cdot \sigma'_i + (1 - \alpha) \cdot \sigma'_e \\
    \end{split}
\end{equation}

This blending mechanism ensures that the final outputs—mean and standard deviation—accurately mirror the non-stationary properties of the data. In order to further adapt to the variation in data distributions, these blended statistics are then processed through linear layers for additional shifting and scaling to make those more robust against the distribution shift:

\begin{equation}
    \begin{split}
        \mu_{\textit{style}} = \text{MLP}_\textit{shift}(\mu_{\textit{blend}}) \\
        \sigma_{\textit{style}} = \text{MLP}_\textit{scale}(\sigma_{\textit{blend}})
    \end{split}
\end{equation}

By adjusting the mean and variance through the Style Blending Gate, the module effectively handles the diverse conditions of non-stationary data. This approach prevents over-stationarization and maintains the essential temporal dynamics that are critical for accurate real-world time series forecasting.

\subsubsection{AdaIN}
\label{subsubsec:adain}

Adaptive Instance Normalization (AdaIN), first popularized by \citet{Huang2017} for its effective use in style transfer, provides a simple yet powerful way to transfer style features—captured by mean and standard deviation—from one input to another. In style transfer, this method involves adapting the visual style of one image—such as color, texture, or patterns—onto the content structure of another, by aligning the statistical properties of the target style with the content image. Although primarily used in style transfer (as formalized in Equation \ref{adain-eq}), AdaIN’s mechanism is highly relevant to our context.

In our model, let $x_{\textit{stream}}$ represent the streaming matrix, with $\mu_{\textit{style}}$ and $\sigma_{\textit{style}}$ serving as the style statistics that guide this matrix. The output $z$ from AdaIN is then given by:

\begin{equation}
    z = \text{AdaIN}(x_{\textit{stream}}, \mu_{\textit{style}}, \sigma_{\textit{style}})
\end{equation}

The function is mathematically expressed as:

\begin{equation}\label{adain-eq}
    \begin{split}
        \text{AdaIN}&(x_{\textit{stream}}, \mu_{\textit{style}}, \sigma_{\textit{style}})=\\
        &\sigma_{\textit{style}}\left( \frac{x_{\textit{stream}}-\mu_{x_{\textit{stream}}}}{\sigma_{x_{\textit{stream}}}}\right) + \mu_{\textit{style}}
    \end{split}
\end{equation}

This integration of AdaIN into our model involves generating a style input that captures the approximations of the non-stationary characteristics through the Style Blending Gate and linear transformations. This style input is then utilized via AdaIN to guide our stationarized input. This approach aims to reintroduce and preserve the intrinsic non-stationary information of the original series, thereby recovering the vital dynamics lost during stationarization.

\subsection{Stacked kronecker product layer}
\label{subsec:skpl}

In this paper, we incorporate the Stacked Kronecker Product Layer, originally introduced as a space efficient and expressive linear layer in the ShapeShifter model \citep{Panahi2021}, to enhance the efficiency and scalability of our architecture. This layer plays a crucial role in reducing the number of trainable parameters, particularly in the large linear layers of our model. During our model development, we observed that concatenating output matrices from CANet layers significantly increased model complexity. However, by leveraging the Kronecker product for matrix factorization, the Stacked Kronecker Product Layer allows us to decompose large weight matrices into smaller, more manageable components while preserving the expressive capacity required for accurate predictions. 

This factorization not only contributes to building a lightweight model but also substantially reduces computational demands without sacrificing performance. Detailed results from our ablation study, presented in Section \ref{sec:model-complexity} and Table \ref{tbl:model-complexity}, validate the effectiveness of this approach. The inclusion of the Stacked Kronecker Product Layer aligns with our goal of designing a streamlined and efficient time series forecasting model, making it a foundational technique that enhances the compactness and performance of CANet.

\begin{table*}[width=1.75\linewidth,cols=5,pos=h]
    \caption{Summary of each dataset. A lower ADF test statistic signifies that the time series data is more stationary.}
    \label{tbl:dataset-summary}
    \begin{tabular*}{\tblwidth}{@{}CCCCC@{}}
        \toprule
        Dataset Name &  Variates & Sampling Frequency & Record Count & ADF\\
        \midrule
        Exchange & 8 & 1 Day & 7,588 & -1.90\\
        ETTh2 & 7 & 1 Hour & 17,420 & -4.13\\
        ILI & 7 & 1 Week & 966 & -5.33\\
        ETTm2 & 7 & 15 Minutes & 69,680 & -5.66\\
        ETTh1 & 7 & 1 Hour & 17,420 & -5.91\\
        Electricity & 321 & 1 Hour & 26,304 & -8.44\\
        ETTm1 & 7 & 15 Minutes & 69,680 & -14.98\\
        Traffic & 862 & 1 Hour & 17,544 & -15.02\\
        Weather & 21 & 10 Minutes & 52,696 & -26.68\\
        \bottomrule
    \end{tabular*}
\end{table*}

\begin{table}
    \caption{Hyperparameter search space. The notation $a \sim b$ denotes a continuous interval from $a$ to $b$, while $\{\dots\}$ represents a discrete set of values.}
    \label{tbl:hyperparam-search-space}
    \begin{tabular*}{\tblwidth}{@{}C|C@{}}
        \toprule
        \multicolumn{1}{c}{Hyperparameter} & \multicolumn{1}{c}{Search Space} \\
        \midrule
        Learning Rate &  $1 \times 10^{-4} \sim 1 \times 10^{-2}$ \\
        Dropout Rate & $0.3 \sim 0.5$ \\
        Batch Size & $\{16, 32, 64, 128, 256, 512, 1024\}$ \\
        Patch Sizes & $\{[8, 32], [8, 48], [8, 64], [16, 48], [16, 64]\}$ \\
        Embedding Dimension & $\{32, 64, 128\}$ \\
        \bottomrule
    \end{tabular*}
\end{table}

\vspace{-1mm}
\subsection{Objective function}
\label{subsec:objective-func}

The squared error, commonly known as L2 loss, is widely used in time series forecasting—including our approach—to measure the squared deviations between predicted values and actual observations over the forecasting period. This loss function is defined as:

\begin{equation}
    \text{Loss}=\text{min}_{t\in \Omega_{\text{Train}}}\sum^N_{j=1}\sum^O_{i=1}||Y^j_{t+i}-\hat{Y}^j_{t+i}||^2_F
\end{equation}

where $\Omega_{\text{Train}}$ represents the training set timestamps, $||\cdot||_F$ the Frobenius norm for matrices, $t$  the current time step, $O$ the forecast horizon, and  $N$ the count of series variables.

\subsection{Evaluation metrics}
\label{subsec:eval-metrics}

In assessing the performance of CANet, this study employs two commonly used evaluation metrics: the mean absolute error (MAE) and the mean squared error (MSE). These metrics are crucial for quantifying the accuracy of the forecasts generated by the model.

The MAE is defined as:

\begin{equation}
    \text{MAE}=\frac{1}{O\times N}\sum^N_{j=1}\sum^O_{i=1}|Y^j_{t+i}-\hat{Y}^j_{t+i}|
\end{equation}

where $|Y^j_{t+i}-\hat{Y}^j_{t+i}|$ represents the absolute error between the actual and the predicted values.

Similarly, the MSE is calculated as:

\begin{equation}
    \text{MSE}=\frac{1}{O\times N}\sum^N_{j=1}\sum^O_{i=1}(Y^j_{t+i}-\hat{Y}^j_{t+i})^2
\end{equation}

Here, $(Y^j_{t+i}-\hat{Y}^j_{t+i})^2$ denotes the squared difference between the actual and the predicted values.

In these equations, $t$ denotes the current time step, $O$ represents the prediction horizon, and $N$ is the number of variables in the predicted time series.

\setlength{\tabcolsep}{4pt}
\begin{table*}[width=2\linewidth,cols=20,pos=h]
    \caption{Multivariate long-term forecasting results. For the ILI dataset, we set the prediction lengths $O \in \{24, 36, 48, 60\}$ with a look-back window of $L=36$, while for the other datasets, the prediction lengths are $O \in \{96, 192, 336, 720\}$ with a look-back window of $L=96$. The best results are highlighted in \textcolor{blue}{blue}, and the second-best in \textcolor{red}{red}.}
    \label{tbl:forecasting-results}
    \scriptsize
    \begin{tabular*}{\tblwidth}{@{}CC|CC|CC|CC|CC|CC|CC|CC|CC|CC@{}}
        \toprule
        \multicolumn{2}{l}{Methods} & \multicolumn{2}{c}{\textbf{CANet}} & \multicolumn{2}{c}{NSTransformer} & \multicolumn{2}{c}{iTransformer} & \multicolumn{2}{c}{PatchTST} & \multicolumn{2}{c}{FEDformer} & \multicolumn{2}{c}{Autoformer} & \multicolumn{2}{c}{TSLANet} & \multicolumn{2}{c}{TimesNet} & \multicolumn{2}{c}{Koopa} \\
        \cmidrule(lr) {3-4} \cmidrule(lr) {5-6} \cmidrule(lr) {7-8} \cmidrule(lr) {9-10} \cmidrule(lr) {11-12} \cmidrule(lr) {13-14} \cmidrule(lr) {15-16} \cmidrule(lr) {17-18} \cmidrule(lr) {19-20}
        \multicolumn{2}{l}{Metric} & \multicolumn{1}{c}{MSE} & \multicolumn{1}{c}{MAE} & \multicolumn{1}{c}{MSE} & \multicolumn{1}{c}{MAE} & \multicolumn{1}{c}{MSE} & \multicolumn{1}{c}{MAE} & \multicolumn{1}{c}{MSE} & \multicolumn{1}{c}{MAE} & \multicolumn{1}{c}{MSE} & \multicolumn{1}{c}{MAE} & \multicolumn{1}{c}{MSE} & \multicolumn{1}{c}{MAE} & \multicolumn{1}{c}{MSE} & \multicolumn{1}{c}{MAE} & \multicolumn{1}{c}{MSE} & \multicolumn{1}{c}{MAE} & \multicolumn{1}{c}{MSE} & \multicolumn{1}{c}{MAE} \\
        \midrule
        \multirow{5}{*}{\begin{turn}{90}\textit{ETTh1}\end{turn}} & 96 & \textcolor{blue}{0.355} & \textcolor{blue}{0.388} & 0.513 & 0.491 & 0.386 & 0.405 & 0.393 & 0.408 & \textcolor{red}{0.376} & 0.419 & 0.452 & 0.460 & 0.387 & 0.405 & 0.384 & \textcolor{red}{0.402} & 0.386 & 0.408 \\
         & 192 & \textcolor{blue}{0.382} & \textcolor{blue}{0.404} & 0.534 & 0.504 & 0.441 & 0.436 & 0.445 & 0.434 & \textcolor{red}{0.420} & 0.448 & 0.467 & 0.464 & 0.448 & 0.436 & 0.436 & \textcolor{red}{0.429} & 0.445 & 0.433 \\
         & 336 & \textcolor{blue}{0.408} & \textcolor{blue}{0.429} & 0.588 & 0.535 & 0.487 & 0.458 & 0.484 & 0.451 & 0.459 & 0.465 & 0.502 & 0.489 & \textcolor{red}{0.451} & \textcolor{red}{0.437} & 0.491 & 0.469 & 0.489 & 0.460 \\
         & 720 & \textcolor{blue}{0.453} & \textcolor{blue}{0.467} & 0.643 & 0.616 & 0.503 & 0.491 & 0.480 & \textcolor{red}{0.471} & 0.506 & 0.507 & \textcolor{red}{0.479} & 0.494 & 0.506 & 0.486 & 0.521 & 0.500 & 0.497 & 0.480 \\
        \cmidrule(lr) {3-20}
         & Avg. & \textcolor{blue}{0.400} & \textcolor{blue}{0.422} & 0.570 & 0.537 & 0.454 & 0.448 & 0.451 & \textcolor{red}{0.441} & \textcolor{red}{0.440} & 0.460 & 0.475 & 0.477 & 0.448 & \textcolor{red}{0.441} & 0.458 & 0.450 & 0.454 & 0.445 \\
        \midrule
        \multirow{5}{*}{\begin{turn}{90}\textit{ETTh2}\end{turn}} & 96 & \textcolor{blue}{0.229} & \textcolor{blue}{0.303} & 0.476 & 0.458 & 0.297 & 0.349 & 0.294 & \textcolor{red}{0.343} & 0.346 & 0.388 & 0.383 & 0.432 & \textcolor{red}{0.289} & 0.345 & 0.340 & 0.374 & 0.320 & 0.361 \\
        & 192 & \textcolor{blue}{0.290} & \textcolor{blue}{0.343} & 0.512 & 0.493 & 0.380 & 0.400 & 0.377 & 0.393 & 0.429 & 0.439 & 0.427 & 0.437 & \textcolor{red}{0.362} & \textcolor{red}{0.391} & 0.402 & 0.414 & 0.378 & 0.398 \\
        & 336 & \textcolor{blue}{0.334} & \textcolor{blue}{0.377} & 0.552 & 0.551 & 0.428 & 0.432 & 0.381 & 0.409 & 0.496 & 0.487 & 0.484 & 0.482 & \textcolor{red}{0.350} & \textcolor{red}{0.389} & 0.452 & 0.452 & 0.415 & 0.429 \\
        & 720 & \textcolor{blue}{0.401} & \textcolor{blue}{0.430} & 0.562 & 0.560 & 0.427 & 0.445 & \textcolor{red}{0.412} & \textcolor{red}{0.433} & 0.463 & 0.474 & 0.474 & 0.485 & 0.418 & 0.439 & 0.462 & 0.468 & 0.446 & 0.456 \\
        \cmidrule(lr) {3-20}
         & Avg. & \textcolor{blue}{0.314} & \textcolor{blue}{0.363} & 0.526 & 0.516 & 0.383 & 0.407 & 0.366 & 0.395 & 0.434 & 0.447 & 0.442 & 0.459 & \textcolor{red}{0.355} & \textcolor{red}{0.391} & 0.414 & 0.427 & 0.390 & 0.411 \\
        \midrule
        \multirow{5}{*}{\begin{turn}{90}\textit{ETTm1}\end{turn}} & 96 & \textcolor{blue}{0.315} & \textcolor{blue}{0.357} & 0.386 & 0.398 & 0.334 & 0.368 & \textcolor{red}{0.321} & \textcolor{red}{0.360} & 0.379 & 0.419 & 0.567 & 0.503 & \textcolor{red}{0.321} & 0.362 & 0.338 & 0.375 & 0.322 & \textcolor{red}{0.360} \\
        & 192 & \textcolor{blue}{0.355} & \textcolor{blue}{0.380} & 0.459 & 0.444 & 0.377 & 0.391 & \textcolor{red}{0.362} & \textcolor{red}{0.384} & 0.426 & 0.441 & 0.591 & 0.514 & \textcolor{red}{0.362} & \textcolor{red}{0.384} & 0.374 & 0.387 & 0.380 & 0.393 \\
        & 336 & \textcolor{blue}{0.371} & \textcolor{blue}{0.401} & 0.495 & 0.464 & 0.426 & 0.420 & 0.392 & \textcolor{red}{0.402} & 0.445 & 0.459 & 0.648 & 0.542 & \textcolor{red}{0.385} & 0.404 & 0.410 & 0.411 & 0.401 & 0.411 \\
        & 720 & \textcolor{blue}{0.431} & \textcolor{blue}{0.430} & 0.585 & 0.516 & 0.491 & 0.459 & 0.450 & \textcolor{red}{0.435} & 0.543 & 0.490 & 0.641 & 0.537 & \textcolor{red}{0.446} & 0.438 & 0.478 & 0.450 & 0.474 & 0.448 \\
        \cmidrule(lr) {3-20}
        & Avg & \textcolor{blue}{0.368} & \textcolor{blue}{0.392} & 0.481 & 0.456 & 0.407 & 0.410 & 0.381 & \textcolor{red}{0.395} & 0.448 & 0.452 & 0.612 & 0.524 & \textcolor{red}{0.379} & 0.397 & 0.400 & 0.406 & 0.394 & 0.403 \\
        \midrule
        \multirow{5}{*}{\begin{turn}{90}\textit{ETTm2}\end{turn}} & 96 & \textcolor{blue}{0.173} & \textcolor{blue}{0.258} & 0.192 & 0.274 & 0.180 & 0.264 & \textcolor{red}{0.178} & \textcolor{red}{0.260} & 0.203 & 0.287 & 0.255 & 0.339 & 0.179 & 0.261 & 0.187 & 0.267 & 0.180 & 0.261 \\
        & 192 & \textcolor{blue}{0.236} & \textcolor{blue}{0.299} & 0.280 & 0.339 & 0.250 & 0.309 & 0.249 & 0.307 & 0.269 & 0.328 & 0.281 & 0.340 & \textcolor{red}{0.243} & \textcolor{red}{0.303} & 0.249 & 0.309 & 0.246 & 0.306 \\
        & 336 & \textcolor{blue}{0.287} & \textcolor{blue}{0.333} & 0.334 & 0.361 & 0.311 & 0.348 & 0.313 & 0.346 & 0.325 & 0.366 & 0.339 & 0.372 & \textcolor{red}{0.308} & \textcolor{red}{0.345} & 0.321 & 0.351 & 0.309 & 0.347 \\
        & 720 & \textcolor{blue}{0.371} & \textcolor{blue}{0.385} & 0.417 & 0.413 & 0.412 & 0.407 & \textcolor{red}{0.400} & \textcolor{red}{0.398} & 0.421 & 0.415 & 0.422 & 0.419 & 0.403 & 0.400 & 0.408 & 0.403 & 0.406 & 0.403 \\
        \cmidrule(lr) {3-20}
        & Avg & \textcolor{blue}{0.267} & \textcolor{blue}{0.319} & 0.306 & 0.347 & 0.288 & 0.332 & 0.285 & 0.328 & 0.305 & 0.349 & 0.324 & 0.368 & \textcolor{red}{0.283} & \textcolor{red}{0.327} & 0.291 & 0.333 & 0.285 & 0.329 \\
        \midrule
        \multirow{5}{*}{\begin{turn}{90}\textit{Electricity}\end{turn}} & 96 & 0.162 & 0.252 & 0.169 & 0.273 & \textcolor{red}{0.148} & \textcolor{blue}{0.240} & 0.174 & 0.259 & 0.193 & 0.308 & 0.201 & 0.317 & 0.171 & 0.256 & 0.168 & 0.272 &  \textcolor{blue}{0.147} & \textcolor{red}{0.247} \\
        & 192 & \textcolor{red}{0.165} & \textcolor{red}{0.262} & 0.182 & 0.286 & \textcolor{blue}{0.162} & \textcolor{blue}{0.253} & 0.178 & 0.265 & 0.201 & 0.315 & 0.222 & 0.334 & 0.178 & 0.263 & 0.184 & 0.289 & 0.181 & 0.276 \\
        & 336 & \textcolor{red}{0.186} & \textcolor{red}{0.279} & 0.200 & 0.304 & \textcolor{blue}{0.178} & \textcolor{blue}{0.269} & 0.196 & 0.282 & 0.214 & 0.329 & 0.231 & 0.338 & 0.195 & 0.280 & 0.198 & 0.300 & 0.195 & 0.290 \\
        & 720 & 0.224 & \textcolor{blue}{0.313} & \textcolor{red}{0.222} & 0.321 & 0.225 & 0.317 & 0.237 & 0.316 & 0.246 & 0.355 & 0.254 & 0.361 & 0.235 & \textcolor{red}{0.314} & \textcolor{blue}{0.220} & 0.320 & 0.229 & 0.316 \\
        \cmidrule(lr) {3-20}
        & Avg & \textcolor{red}{0.184} & \textcolor{red}{0.277} & 0.193 & 0.296 & \textcolor{blue}{0.178} & \textcolor{blue}{0.270} & 0.196 & 0.281 & 0.214 & 0.327 & 0.227 & 0.338 & 0.195 & 0.278 & 0.193 & 0.295 & 0.188 & 0.282 \\
        \midrule
        \multirow{5}{*}{\begin{turn}{90}\textit{Exchange}\end{turn}} & 96 & \textcolor{blue}{0.078} & \textcolor{blue}{0.197} & 0.111 & 0.237 & 0.086 & 0.206 & \textcolor{red}{0.081} & \textcolor{red}{0.198} & 0.148 & 0.278 & 0.197 & 0.323 & 0.083 & 0.201 & 0.107 & 0.234 & 0.087 & 0.207 \\
        & 192 & \textcolor{blue}{0.157} & \textcolor{blue}{0.287} & 0.219 & 0.335 & 0.177 & 0.299 & \textcolor{red}{0.172} & \textcolor{red}{0.294} & 0.271 & 0.315 & 0.300 & 0.369 & 0.177 & 0.299 & 0.226 & 0.344 & 0.182 & 0.305 \\
        & 336 & \textcolor{blue}{0.276} & \textcolor{blue}{0.386} & 0.421 & 0.476 & 0.331 & 0.417 & \textcolor{red}{0.320} & \textcolor{red}{0.408} & 0.460 & 0.427 & 0.509 & 0.524 & 0.331 & 0.417 & 0.367 & 0.448 & 0.356 & 0.434 \\
        & 720 & \textcolor{blue}{0.490} & \textcolor{blue}{0.539} & 1.092 & 0.769 & \textcolor{red}{0.847} & \textcolor{red}{0.691} & 0.864 & 0.696 & 1.195 & 0.695 & 1.447 & 0.941 & 0.888 & 0.739 & 0.964 & 0.746 & 0.953 & 0.742 \\
        \cmidrule(lr) {3-20}
        & Avg & \textcolor{blue}{0.250} & \textcolor{blue}{0.352} & 0.461 & 0.454 & 0.360 & 0.403 & \textcolor{red}{0.359} & \textcolor{red}{0.399} & 0.519 & 0.429 & 0.613 & 0.539 & 0.370 & 0.414 & 0.416 & 0.443 & 0.395 & 0.422 \\
        \midrule
        \multirow{5}{*}{\begin{turn}{90}\textit{Weather}\end{turn}} & 96 & \textcolor{red}{0.161} & \textcolor{red}{0.211} & 0.173 & 0.223 & 0.174 & 0.214 & 0.178 & 0.219 & 0.217 & 0.296 & 0.266 & 0.336 & 0.176 & 0.215 & 0.172 & 0.220 & \textcolor{blue}{0.159} & \textcolor{blue}{0.204} \\
        & 192 & \textcolor{blue}{0.204} & \textcolor{blue}{0.251} & 0.245 & 0.285 & 0.221 & \textcolor{red}{0.254} & 0.224 & 0.259 & 0.276 & 0.336 & 0.307 & 0.367 & 0.225 & 0.258 & 0.219 & 0.261 & \textcolor{red}{0.209} & \textcolor{blue}{0.251} \\
        & 336 & \textcolor{blue}{0.245} & \textcolor{blue}{0.288} & 0.321 & 0.338 & 0.278 & 0.296 & 0.278 & 0.298 & 0.339 & 0.380 & 0.359 & 0.395 & 0.279 & 0.297 & 0.280 & 0.306 & \textcolor{red}{0.266} & \textcolor{red}{0.291} \\
        & 720 & \textcolor{blue}{0.321} & \textcolor{blue}{0.339} & 0.414 & 0.410 & 0.358 & 0.347 & 0.350 & \textcolor{red}{0.346} & 0.403 & 0.428 & 0.419 & 0.428 & 0.355 & \textcolor{red}{0.346} & 0.365 & 0.359 & \textcolor{red}{0.348} & \textcolor{red}{0.346} \\
        \cmidrule(lr) {3-20}
        & Avg & \textcolor{blue}{0.233} & \textcolor{blue}{0.272} & 0.288 & 0.314 & 0.258 & 0.278 & 0.258 & 0.281 & 0.309 & 0.360 & 0.338 & 0.382 & 0.259 & 0.279 & 0.259 & 0.287 & \textcolor{red}{0.246} & \textcolor{red}{0.273} \\
        \midrule
        \multirow{5}{*}{\begin{turn}{90}\textit{ILI}\end{turn}} & 24 & \textcolor{red}{1.497} & \textcolor{red}{0.765} & 2.294 & 0.945 & 2.600 & 1.046 & \textcolor{blue}{1.281} & \textcolor{blue}{0.704} & 3.228 & 1.260 & 3.483 & 1.287 & 2.521 & 1.002 & 2.317 & 0.934 & 1.870 & 0.877 \\
        & 36 & \textcolor{red}{1.461} & \textcolor{blue}{0.748} & 1.825 & 0.848 & 2.481 & 1.025 & \textcolor{blue}{1.251} & \textcolor{red}{0.752} & 2.679 & 1.080 & 3.103 & 1.148 & 2.074 & 0.926 & 1.972 & 0.920 & 1.968 & 0.925 \\
        & 48 & \textcolor{blue}{1.461} & \textcolor{blue}{0.778} & 2.010 & 0.900 & 2.499 & 1.034 & \textcolor{red}{1.901} & \textcolor{red}{0.879} & 2.622 & 1.078 & 2.669 & 1.085 & 2.333 & 0.975 & 2.238 & 0.940 & 1.904 & 0.924 \\
        & 60 & \textcolor{red}{1.863} & \textcolor{red}{0.888} & 2.178 & 0.963 & 2.629 & 1.067 & \textcolor{blue}{1.611} & \textcolor{blue}{0.844} & 2.857 & 1.157 & 2.770 & 1.125 & 2.119 & 0.955 & 2.027 & 0.928 & 1.884 & 0.892 \\
        \cmidrule(lr) {3-20}
        & Avg & \textcolor{red}{1.571} & \textcolor{blue}{0.795} & 2.077 & 0.914 & 2.552 & 1.043 & \textcolor{blue}{1.511} & \textcolor{blue}{0.795} & 2.847 & 1.144 & 3.006 & 1.161 & 2.262 & 0.965 & 2.139 & 0.931 & 1.907 & \textcolor{red}{0.905} \\
        \midrule
        \multirow{5}{*}{\begin{turn}{90}\textit{Traffic}\end{turn}} & 96 & 0.522 & 0.333 & 0.612 & 0.338 & \textcolor{blue}{0.395} & \textcolor{blue}{0.268} & 0.477 & \textcolor{red}{0.305} & 0.587 & 0.366 & 0.613 & 0.388 & \textcolor{red}{0.475} & 0.307 & 0.593 & 0.321 & 0.477 & 0.316 \\
        & 192 & 0.505 & 0.322 & 0.613 & 0.340 & \textcolor{blue}{0.417} & \textcolor{blue}{0.276} & \textcolor{red}{0.471} & \textcolor{red}{0.299} & 0.604 & 0.373 & 0.616 & 0.382 & 0.478 & 0.306 & 0.617 & 0.336 & 0.501 & 0.340 \\
        & 336 & 0.532 & 0.338 & 0.618 & 0.328 & \textcolor{blue}{0.433} & \textcolor{blue}{0.283} & \textcolor{red}{0.485} & \textcolor{red}{0.305} & 0.621 & 0.383 & 0.622 & 0.337 & 0.494 & 0.312 & 0.629 & 0.336 & 0.531 & 0.349 \\
        & 720 & 0.554 & 0.352 & 0.653 & 0.355 & \textcolor{blue}{0.467} & \textcolor{blue}{0.302} & \textcolor{red}{0.518} & \textcolor{red}{0.325} & 0.626 & 0.382 & 0.660 & 0.408 & 0.528 & 0.468 & 0.640 & 0.350 & 0.566 & 0.366 \\
        \cmidrule(lr) {3-20}
        & Avg & 0.528 & 0.336 & 0.624 & 0.340 & \textcolor{blue}{0.428} & \textcolor{blue}{0.282} & \textcolor{red}{0.488} & \textcolor{red}{0.309} & 0.610 & 0.376 & 0.628 & 0.379 & 0.494 & 0.348 & 0.620 & 0.336 & 0.519 & 0.343 \\
        \midrule
        \multicolumn{2}{l}{Count} & \multicolumn{2}{c}{50} & \multicolumn{2}{c}{0} & \multicolumn{2}{c}{13} & \multicolumn{2}{c}{5} & \multicolumn{2}{c}{0} & \multicolumn{2}{c}{0} & \multicolumn{2}{c}{0} & \multicolumn{2}{c}{1} & \multicolumn{2}{c}{4} \\
        \bottomrule
    \end{tabular*}
\end{table*}

\setlength{\tabcolsep}{4pt}
\begin{table*}[width=2\linewidth,cols=20,pos=h]
    \caption{Impact of various normalization methods.  
    LN, BN, IN denote Layer, Batch, Instance, Normalizations, respectively.  
    We set the look-back window $L=96$ and prediction lengths $O\in\{96,192,336,720\}$.  
    \textcolor{blue}{Blue} highlights the best result, while \textcolor{red}{red} marks the second-best result.}
    \label{tbl:norm-comparison}
    \scriptsize
    \begin{tabular*}{\tblwidth}{@{}CC|CC|CC|CC|CC|CC|CC|CC|CC|CC@{}}
        \toprule
        \multicolumn{2}{l}{Methods} 
            & \multicolumn{2}{c}{\textbf{NSAN}} 
            & \multicolumn{2}{c}{LN} 
            & \multicolumn{2}{c}{BN} 
            & \multicolumn{2}{c}{IN} 
            & \multicolumn{2}{c}{DAIN}
            & \multicolumn{2}{c}{RevIN} 
            & \multicolumn{2}{c}{SAN} 
            & \multicolumn{2}{c}{FAN} 
            & \multicolumn{2}{c}{Dish-TS} \\
        \cmidrule(lr){3-4}\cmidrule(lr){5-6}\cmidrule(lr){7-8}\cmidrule(lr){9-10}
        \cmidrule(lr){11-12}\cmidrule(lr){13-14}\cmidrule(lr){15-16}\cmidrule(lr){17-18}\cmidrule(lr){19-20}
        \multicolumn{2}{l}{Metric} 
            & MSE & MAE & MSE & MAE & MSE & MAE & MSE & MAE & MSE & MAE
            & MSE & MAE & MSE & MAE & MSE & MAE & MSE & MAE \\
        \midrule
        \multirow{5}{*}{\begin{turn}{90}\textit{Exchange}\end{turn}}
            & 96  & \textcolor{red}{0.078} & \textcolor{blue}{0.197} & 0.080 & \textcolor{red}{0.198} & 0.080 & \textcolor{red}{0.198} & 0.080 & \textcolor{red}{0.198} & 0.079 & \textcolor{blue}{0.197} & 0.080 & 0.199 & \textcolor{blue}{0.077} & 0.203 & 0.084 & 0.212 & 0.086 & 0.206 \\
            & 192 & 0.157 & \textcolor{blue}{0.287} & 0.168 & 0.292 & 0.166 & 0.290 & 0.168 & 0.292 & 0.168 & 0.294 & 0.182 & 0.319 & \textcolor{blue}{0.150} & \textcolor{red}{0.288} & 0.170 & 0.308 & \textcolor{red}{0.154} & 0.295 \\
            & 336 & 0.276 & \textcolor{red}{0.386} & 0.319 & 0.407 & 0.300 & 0.397 & 0.297 & 0.395 & 0.320 & 0.408 & 0.345 & 0.436 & \textcolor{red}{0.250} & 0.388 & 0.345 & 0.437 & \textcolor{blue}{0.233} & \textcolor{blue}{0.382} \\
            & 720 & \textcolor{red}{0.490} & \textcolor{blue}{0.539} & 0.795 & 0.668 & 0.811 & 0.676 & 0.795 & 0.668 & 0.816 & 0.679 & 0.805 & 0.672 & 0.604 & 0.624 & 0.646 & 0.626 & \textcolor{blue}{0.483} & \textcolor{red}{0.578} \\
        \cmidrule(lr){2-20}
            & Avg.& \textcolor{blue}{0.250} & \textcolor{red}{0.352} & 0.341 & 0.391 & 0.339 & 0.390 & 0.335 & 0.388 & 0.346 & 0.395 & 0.353 & 0.407 & 0.270 & 0.376 & 0.311 & 0.396 & \textcolor{blue}{0.239} & \textcolor{red}{0.365} \\
        \midrule
        \multirow{5}{*}{\begin{turn}{90}\textit{ETTh1}\end{turn}}
            & 96  & \textcolor{blue}{0.355} & \textcolor{blue}{0.388} & \textcolor{red}{0.360} & \textcolor{red}{0.391} & 0.362 & 0.392 & 0.362 & 0.392 & 0.362 & 0.394 & 0.362 & 0.393 & 0.376 & 0.395 & 0.386 & 0.406 & 0.381 & 0.405 \\
            & 192 & \textcolor{blue}{0.382} & \textcolor{blue}{0.404} & 0.389 & 0.409 & 0.389 & 0.412 & \textcolor{red}{0.386} & 0.409 & 0.387 & 0.410 & \textcolor{red}{0.386} & \textcolor{red}{0.408} & 0.419 & 0.413 & 0.427 & 0.422 & 0.423 & 0.421 \\
            & 336 & \textcolor{red}{0.408} & \textcolor{blue}{0.429} & 0.412 & 0.434 & 0.412 & \textcolor{blue}{0.429} & \textcolor{blue}{0.407} & \textcolor{red}{0.431} & 0.417 & \textcolor{blue}{0.429} & 0.414 & \textcolor{blue}{0.429} & 0.441 & 0.435 & 0.469 & 0.452 & 0.453 & 0.456 \\
            & 720 & \textcolor{red}{0.453} & 0.467 & 0.454 & \textcolor{blue}{0.461} & 0.464 & \textcolor{blue}{0.461} & 0.454 & \textcolor{blue}{0.461} & 0.470 & \textcolor{red}{0.462} & \textcolor{blue}{0.449} & 0.464 & 0.535 & 0.517 & 0.526 & 0.500 & 0.528 & 0.507 \\
        \cmidrule(lr){2-20}
            & Avg.& \textcolor{blue}{0.400} & \textcolor{blue}{0.422} & 0.404 & 0.424 & 0.407 & 0.424 & \textcolor{red}{0.402} & \textcolor{red}{0.423} & 0.409 & 0.424 & 0.403 & 0.424 & 0.443 & 0.440 & 0.452 & 0.445 & 0.446 & 0.447 \\
        \midrule
        \multirow{5}{*}{\begin{turn}{90}\textit{ETTh2}\end{turn}}
            & 96  & 0.229 & \textcolor{red}{0.303} & \textcolor{red}{0.228} & 0.304 & 0.230 & 0.304 & \textcolor{blue}{0.227} & 0.304 & \textcolor{red}{0.228} & \textcolor{blue}{0.302} & \textcolor{red}{0.228} & 0.304 & 0.251 & 0.327 & 0.243 & 0.323 & 0.242 & 0.321 \\
            & 192 & \textcolor{blue}{0.290} & \textcolor{red}{0.343} & \textcolor{red}{0.291} & \textcolor{red}{0.343} & \textcolor{red}{0.291} & \textcolor{red}{0.343} & \textcolor{red}{0.291} & \textcolor{red}{0.343} & \textcolor{blue}{0.290} & \textcolor{blue}{0.342} & \textcolor{blue}{0.290} & \textcolor{red}{0.343} & 0.315 & 0.371 & 0.306 & 0.367 & 0.303 & 0.353 \\
            & 336 & \textcolor{red}{0.334} & \textcolor{blue}{0.377} & 0.341 & 0.382 & 0.340 & \textcolor{red}{0.380} & 0.341 & 0.381 & 0.341 & 0.381 & 0.342 & 0.382 & 0.362 & 0.409 & \textcolor{blue}{0.333} & 0.387 & 0.363 & 0.404 \\
            & 720 & \textcolor{blue}{0.401} & \textcolor{red}{0.430} & 0.407 & \textcolor{red}{0.430} & \textcolor{red}{0.405} & \textcolor{blue}{0.429} & 0.407 & \textcolor{red}{0.430} & \textcolor{red}{0.405} & \textcolor{blue}{0.429} & 0.407 & \textcolor{red}{0.430} & 0.435 & 0.455 & 0.540 & 0.513 & 0.486 & 0.482 \\
        \cmidrule(lr){2-20}
            & Avg.& \textcolor{blue}{0.314} & \textcolor{blue}{0.363} & 0.317 & 0.365 & 0.317 & \textcolor{red}{0.364} & 0.317 & 0.365 & \textcolor{red}{0.316} & \textcolor{red}{0.364} & 0.317 & 0.365 & 0.341 & 0.391 & 0.356 & 0.398 & 0.349 & 0.390 \\
        \midrule
        \multirow{5}{*}{\begin{turn}{90}\textit{ETTm1}\end{turn}}
            & 96  & \textcolor{blue}{0.315} & \textcolor{blue}{0.357} & 0.323 & 0.362 & 0.325 & 0.363 & \textcolor{red}{0.317} & \textcolor{red}{0.359} & 0.324 & 0.363 & \textcolor{red}{0.317} & \textcolor{red}{0.359} & \textcolor{blue}{0.315} & 0.370 & 0.322 & 0.361 & 0.325 & 0.366 \\
            & 192 & \textcolor{blue}{0.355} & \textcolor{blue}{0.380} & 0.366 & 0.387 & 0.370 & 0.388 & 0.360 & 0.384 & 0.369 & 0.390 & \textcolor{red}{0.357} & \textcolor{red}{0.383} & \textcolor{red}{0.357} & 0.393 & 0.365 & 0.387 & 0.360 & 0.388 \\
            & 336 & \textcolor{blue}{0.371} & \textcolor{blue}{0.401} & 0.392 & 0.407 & 0.383 & \textcolor{red}{0.406} & 0.385 & 0.409 & 0.388 & 0.410 & \textcolor{red}{0.381} & 0.407 & 0.385 & 0.414 & 0.392 & 0.409 & 0.391 & 0.413 \\
            & 720 & \textcolor{blue}{0.431} & \textcolor{blue}{0.430} & 0.446 & 0.442 & 0.443 & 0.441 & 0.444 & 0.440 & 0.443 & 0.440 & \textcolor{red}{0.442} & \textcolor{red}{0.439} & 0.447 & 0.447 & 0.461 & 0.444 & \textcolor{red}{0.442} & 0.442 \\
        \cmidrule(lr){2-20}
            & Avg.& \textcolor{blue}{0.368} & \textcolor{blue}{0.392} & 0.382 & 0.400 & 0.380 & 0.400 & 0.377 & 0.398 & 0.381 & 0.401 & \textcolor{red}{0.374} & \textcolor{red}{0.397} & 0.376 & 0.406 & 0.385 & 0.400 & 0.380 & 0.402 \\
        \midrule
        \multirow{5}{*}{\begin{turn}{90}\textit{ETTm2}\end{turn}}
            & 96  & \textcolor{blue}{0.173} & \textcolor{red}{0.258} & 0.176 & \textcolor{red}{0.258} & 0.176 & 0.259 & 0.176 & \textcolor{blue}{0.258} & 0.176 & \textcolor{red}{0.258} & \textcolor{red}{0.175} & \textcolor{blue}{0.257} & 0.184 & 0.276 & 0.179 & 0.268 & 0.196 & 0.283 \\
            & 192 & \textcolor{blue}{0.236} & \textcolor{blue}{0.299} & 0.242 & 0.302 & \textcolor{red}{0.241} & \textcolor{red}{0.301} & 0.242 & 0.303 & \textcolor{red}{0.241} & \textcolor{red}{0.301} & \textcolor{red}{0.241} & 0.303 & 0.247 & 0.318 & 0.242 & 0.310 & 0.264 & 0.331 \\
            & 336 & \textcolor{blue}{0.287} & \textcolor{blue}{0.333} & 0.294 & 0.339 & \textcolor{red}{0.290} & \textcolor{red}{0.336} & 0.297 & 0.341 & 0.292 & \textcolor{red}{0.336} & 0.293 & 0.339 & 0.313 & 0.361 & 0.304 & 0.349 & 0.327 & 0.372 \\
            & 720 & \textcolor{blue}{0.371} & \textcolor{blue}{0.385} & 0.378 & 0.391 & \textcolor{red}{0.375} & 0.390 & 0.378 & 0.391 & 0.377 & \textcolor{red}{0.389} & 0.380 & 0.392 & 0.396 & 0.416 & 0.395 & 0.408 & 0.457 & 0.455 \\
        \cmidrule(lr){2-20}
            & Avg.& \textcolor{blue}{0.267} & \textcolor{blue}{0.319} & 0.273 & 0.323 & \textcolor{red}{0.271} & 0.322 & 0.273 & 0.323 & 0.272 & \textcolor{red}{0.321} & 0.272 & 0.323 & 0.285 & 0.343 & 0.280 & 0.334 & 0.311 & 0.360 \\
        \midrule
        \multicolumn{2}{l}{Count} 
            & \multicolumn{2}{c}{26} 
            & \multicolumn{2}{c}{1} 
            & \multicolumn{2}{c}{3} 
            & \multicolumn{2}{c}{4} 
            & \multicolumn{2}{c}{6} 
            & \multicolumn{2}{c}{4} 
            & \multicolumn{2}{c}{3} 
            & \multicolumn{2}{c}{1} 
            & \multicolumn{2}{c}{3} \\
        \bottomrule
    \end{tabular*}
\end{table*}

\section{Experiments}
\label{sec:experiments}

In this paper, we conducted comprehensive experiments to evaluate the performance and explore the capabilities of CANet across nine real-world benchmarks for long-term time series forecasting. Additionally, we aimed to confirm the versatility of the proposed framework by testing it on various popular model variants.

\subsection{Datasets}
\label{subsec:datasets}

This section provides a brief overview of each dataset used in the experiments conducted in this paper:

\begin{enumerate}[(1)]
    \item Exchange \citep{Lai2018} contains daily exchange rate panel data for eight countries, covering the period from 1990 to 2016.
    \item Electricity \citep{Zhou2021} consists of hourly electricity consumption data for 321 customers, spanning from 2012 to 2014.
    \item Traffic \citep{Zeng2023} includes hourly road occupancy rates recorded by 862 sensors on highways in the San Francisco Bay Area, covering the time frame from January 2015 to December 2016.
    \item Weather \citep{Zhou2022} comprises meteorological time series data with 21 features, collected every 10 minutes throughout 2020 by the Max Planck Institute for Biogeochemistry.
    \item ILI \citep{Wu2021} tracks the ratio of reported influenza-like illness (ILI) cases to total cases, reported weekly by the U.S. Centers for Disease Control and Prevention, from 2002 to 2021.
    \item ETT \citep{Liu2022} contains four sub-datasets—ETTh1, ETTh2, ETTm1, and ETTm2. These datasets include both hourly and minute-level data from power transformers, spanning from July 2016 to July 2018, capturing the impact of power load and stabilization factors.
\end{enumerate}

This study utilizes the Augmented Dickey-Fuller (ADF) test statistic \citep{Elliott1996} to assess the stationarity of each dataset. The datasets are listed in Table \ref{tbl:dataset-summary} in ascending order of their ADF scores. As a standard practice, each dataset is split into training, validation, and test sets in chronological order, using a 7:1:2 ratio.

\subsection{Baselines}
\label{subsec:baselines}

To evaluate CANet's performance against the current state-of-the-art models, we compared it with seven leading models for time series forecasting. Among Transformer-based models, we included Non-stationary Transformer (NSTransformer) \citep{Liu2022}, Koopa \citep{liu2023koopa}, iTransformer \citep{liu2024itransformer}, PatchTST \citep{Nie2023}, FEDformer \citep{Zhou2022}, and Autoformer \citep{Wu2021}. For CNN-based models, we included TSLANet \citep{Eldele2024} and TimesNet \citep{Wu2023}. When comparing against these baselines, we used their best reported results if they aligned with our experimental settings; otherwise, we re-ran their implementations.

\subsection{Experimental settings}
\label{subsec:experimental-settings}

CANet was developed using the PyTorch deep learning framework \citep{Paszke2019} and trained on four NVIDIA A100 GPUs, each with 80 GB of memory. In order to ensure the reproducibility of our results, we set a fixed random seed throughout the experiments. In order to optimize the model's performance, we employed Optuna \citep{Akiba2019}, a specialized framework for hyperparameter tuning, defining a search space, reported in Table \ref{tbl:hyperparam-search-space}, and conducting 100 training runs to find the best configuration\footnote[2]{Due to time limitations, we limited the Electricity and Traffic datasets with a 720-step prediction length to 25 trials each.}. Details on our approach to hyperparameter optimization are provided in Section \ref{subsec:hyperparam-optim}. In all real-world experiments, we used the past 96 time steps as the look-back window $L$ (except in Section \ref{subsec:varying-lookback-window}). This window was used to predict the subsequent 96, 192, 336, and 720 time steps, following the approach of prior studies \citep{Wu2021, Zhou2022}. The results are evaluated using the average MSE and MAE for clarity.

\subsection{Hyperparameter optimization}
\label{subsec:hyperparam-optim}

To fully leverage CANet’s forecasting potential, we performed extensive hyperparameter tuning using Optuna across all experiments in this study. Our goal was to identify the optimal combination of hyperparameters to maximize our model accuracy. We focused on essential parameters outlined in Table \ref{tbl:hyperparam-search-space}, specifying a search space for each one and performing 100 trials as suggested by \citet{Akiba2019}. In this table, "Patch Sizes" refers to the patch sizes used at each layer of the model, while "Embedding Dimension" indicates the dimension of each patch. We utilized the tree-structured Parzen estimator (TPE), a Bayesian optimization approach that enables efficient exploration by balancing exploration and exploitation of the hyperparameter space, and set a fixed random seed to make our Optuna studies reproducible.

\begin{figure*}[]
  \centering
    \includegraphics[width=0.84\linewidth]{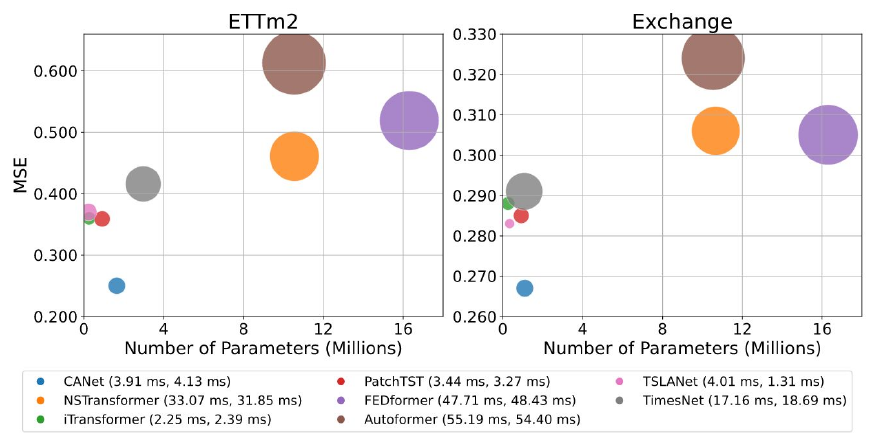}
    \caption[]{Comparison of model complexity on the ETTm2 and Exchange datasets. Each model was tested with prediction length $O \in \{96, 192, 336, 720\}$, with the results averaged across these intervals.}
    \vspace{-6mm}
    \label{fig:model-complexity}
\end{figure*}

\subsection{Long-term multivariate time series forecasting performance}
\label{subsec:long-term-multivariate-time-series-forecasting}

\begin{figure*}[]
    \centering
    \includegraphics[width=0.9\linewidth]{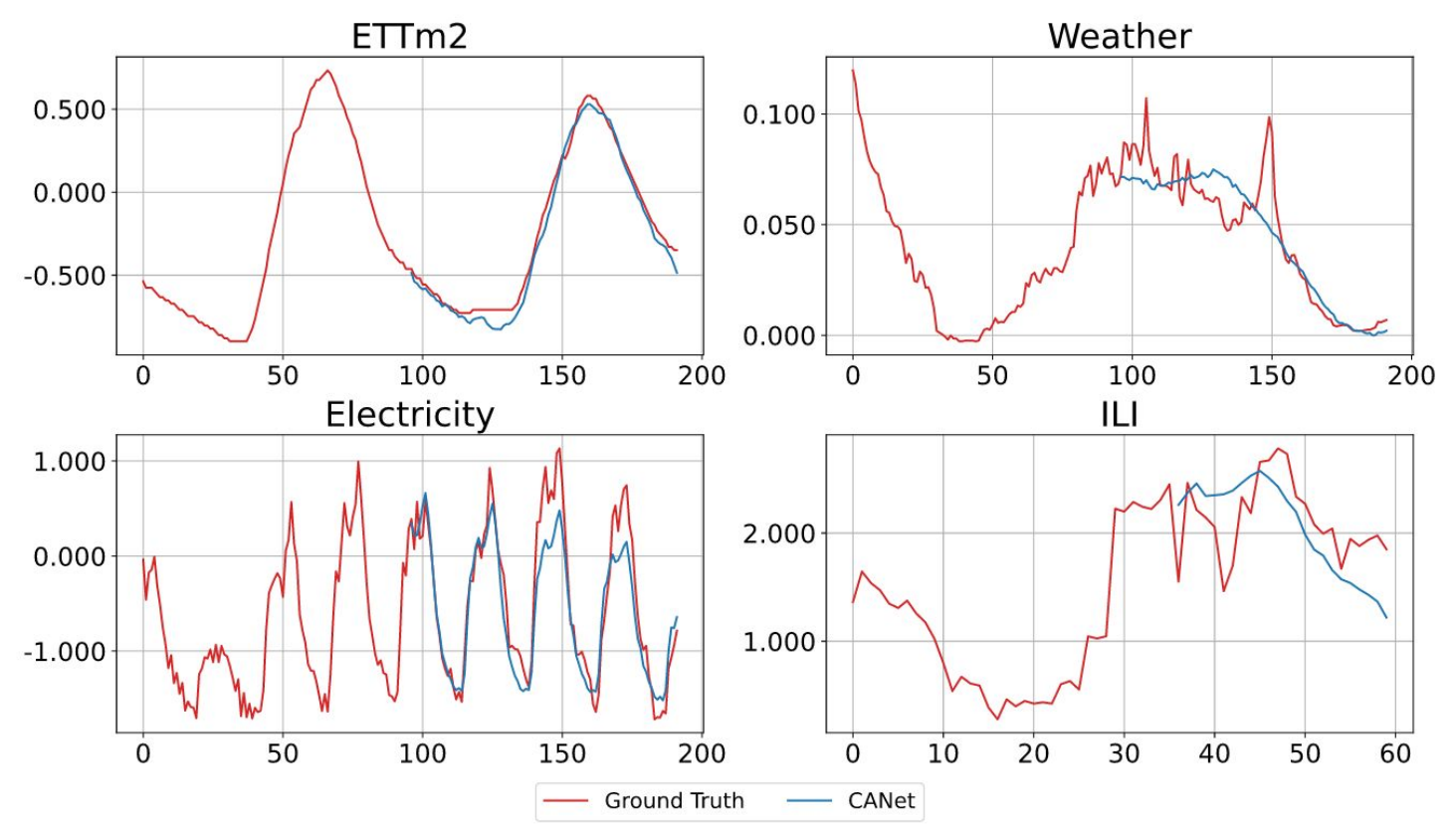}
    \caption[]{Visualization of CANet’s forecasting performance on the ETTm2, Weather, Electricity, and ILI datasets. For the ILI dataset, we set the prediction length $O=24$ with a look-back window of $L=36$, while for the other datasets, the prediction length $O$ is set as 96 with a look-back window of $L=96$.}
    \vspace{-6mm}
    \label{fig:model-pred}
\end{figure*}

CANet exhibits exceptional forecasting capabilities when compared to other models, as shown in Table \ref{tbl:forecasting-results}. Its performance is further illustrated in Figure \ref{fig:model-pred}. Out of 72 forecasting scenarios across nine benchmark datasets, CANet achieved top performance in 50 cases and ranked second in 11 cases. Achieving state-of-the-art results across \textbf{70\%} of scenarios and for various prediction horizons in long-term multivariate forecasting, CANet consistently outperforms leading models described in Section \ref{subsec:baselines}. Its advantages are particularly notable on datasets with high non-stationarity: for a prediction length of 720, CANet reduces MSE by \textbf{42\%} (from $0.847$ to $0.490$) and MAE by \textbf{22\%} (from $0.691$ to $0.539$) in the Exchange dataset. Similarly, for a prediction length of 96, CANet achieves a \textbf{42\%} MSE reduction (from $0.289$ to $0.229$) and a \textbf{22\%} MAE reduction (from $0.343$ to $0.303$) in the ETTh2 dataset compared to previous state-of-the-art models. These results underscore CANet's superiority on non-stationary data and highlight the limitations of other deep models when handling such data intricacies.

\subsection{Effectiveness of Non-stationary Adaptive Normalization}
\label{subsec:nsan-effectiveness}

We investigated the effectiveness of the NSAN module in integrating non-stationary factors into the model. Table \ref{tbl:norm-comparison} presents our findings, comparing it with both widely used normalization techniques such as Layer Normalization \citep{Ba2016}, Batch Normalization \citep{Ioffe2015}, Instance Normalization \citep{Ulyanov2014}, and Deep Adaptive Input Normalization (DAIN) \citep{Passalis2020}, as well as model-agnostic methods, such as RevIN \citep{Kim2022}, SAN \citep{Liu2023}, FAN \citep{ye2024}, and Dish-TS \citep{fan2023dishtsgeneralparadigmalleviating}. To ensure a fair comparison, we substituted only the NSAN module, with each of these normalization techniques within CANet. For model-agnostic methods that are not suitable for direct integration between internal model blocks, we applied them at the model's input/output level by replacing NSAN with a Layer Normalization block, following the approach used in TSLANet. The ETT and Exchange datasets were selected for this evaluation due to their strong non-stationarity. As shown in Table \ref{tbl:norm-comparison}, the NSAN module proves effective in capturing and adapting to the data’s inherent non-stationary characteristics, outperforming both widely adopted and model-agnostic normalization techniques.

\begin{table*}[width=2\linewidth,cols=9]
    \caption{Ablation study and cross-architecture comparison of CANet. We compare the forecasting performance of CANet against its ablated variants by removing Adaptive Spectral Block (ASB), Interactive Convolutional Block (ICB), Multi-resolution Patching (MRP), or Blending Gate (BG), as well as TSLANet (T) integrated with RevIN, AdaIN, and NSAN. All models are evaluated at four prediction horizons ($O \in \{96, 192, 336, 720\}$). \textbf{Bold} highlights the best score.}
    \label{tbl:ablation}
    \begin{tabular*}{\tblwidth}{@{}CC|CC|CC|CC|CC|CC|CC|CC|CC@{}}
        \toprule
        \multicolumn{2}{l}{Dataset} & \multicolumn{2}{c}{CANet} & \multicolumn{2}{c}{w/o ASB} & \multicolumn{2}{c}{w/o ICB} & \multicolumn{2}{c}{w/o MRP} & \multicolumn{2}{c}{w/o BG} & \multicolumn{2}{c}{T w/ RevIN} & \multicolumn{2}{c}{T w/ AdaIN} & \multicolumn{2}{c}{T w/ NSAN} \\
        \cmidrule(lr) {3-4} \cmidrule(lr) {5-6} \cmidrule(lr) {7-8} \cmidrule(lr) {9-10}
        \cmidrule(lr) {11-12} \cmidrule(lr) {13-14} \cmidrule(lr) {15-16} \cmidrule(lr) {17-18}
        \multicolumn{2}{l}{Metric} & MSE & MAE & MSE & MAE & MSE & MAE & MSE & MAE & MSE & MAE & MSE & MAE & MSE & MAE & MSE & MAE \\
        \midrule
\multirow{5}{*}{\begin{turn}{90}\textit{ETTm2}\end{turn}} & 96 & \textbf{0.173} & 0.258 & 0.174 & 0.258 & 0.175 & 0.258 & 0.174 & 0.257 & \textbf{0.173} & \textbf{0.256} & 0.179 & 0.261 & 0.181 & 0.267 & 0.189 & 0.273 \\
  & 192 & \textbf{0.236} & \textbf{0.299} & 0.237 & 0.300 & 0.238 & 0.301 & 0.238 & 0.300 & 0.240 & 0.304 & 0.243 & 0.303 & 0.271 & 0.325 & 0.252 & 0.314 \\
  & 336 & 0.287 & \textbf{0.333} & \textbf{0.286} & 0.334 & 0.287 & \textbf{0.333} & 0.287 & \textbf{0.333} & 0.298 & 0.339 & 0.308 & 0.345 & 0.363 & 0.376 & 0.306 & 0.347 \\
  & 720 & 0.371 & \textbf{0.385} & \textbf{0.370} & 0.387 & 0.371 & 0.386 & \textbf{0.370} & \textbf{0.385} & 0.395 & 0.403 & 0.402 & 0.400 & 0.463 & 0.433 & 0.391 & 0.395 \\
\cmidrule(lr) {2-18}
  & Avg. & \textbf{0.267} & \textbf{0.319} & \textbf{0.267} & 0.320 & 0.268 & 0.320 & \textbf{0.267} & \textbf{0.319} & 0.277 & 0.326 & 0.283 & 0.327 & 0.320 & 0.350 & 0.284 & 0.332 \\
\midrule
\multirow{5}{*}{\begin{turn}{90}\textit{Exchange}\end{turn}} & 96 & \textbf{0.078} & \textbf{0.197} & \textbf{0.078} & \textbf{0.197} & 0.079 & \textbf{0.197} & \textbf{0.078} & \textbf{0.197} & 0.079 & 0.198 & 0.083 & 0.201 & 0.118 & 0.249 & 0.091 & 0.214 \\
  & 192 & \textbf{0.157} & \textbf{0.287} & 0.158 & \textbf{0.287} & 0.158 & \textbf{0.287} & 0.162 & 0.291 & 0.162 & 0.294 & 0.173 & 0.296 & 0.209 & 0.335 & 0.186 & 0.309 \\
  & 336 & 0.276 & 0.386 & 0.273 & 0.387 & \textbf{0.268} & \textbf{0.384} & 0.278 & 0.386 & 0.291 & 0.399 & 0.334 & 0.418 & 0.338 & 0.429 & 0.329 & 0.416 \\
  & 720 & \textbf{0.490} & 0.539 & 0.515 & 0.551 & 0.499 & 0.545 & 0.512 & 0.550 & 0.492 & \textbf{0.532} & 1.070 & 0.779 & 0.530 & 0.560 & 0.720 & 0.638 \\
\cmidrule(lr) {2-18}
  & Avg. & \textbf{0.250} & \textbf{0.352} & 0.256 & 0.356 & 0.251 & 0.353 & 0.258 & 0.356 & 0.252 & 0.356 & 0.415 & 0.424 & 0.299 & 0.393 & 0.332 & 0.394 \\
        \bottomrule
    \end{tabular*}
\end{table*}

\begin{table}[width=0.9\linewidth,cols=6]
    \caption{Impact of the Stacked Kronecker Product Layer (SKPL) on CANet’s complexity and performance. The look-back window $L$ is set to 96, with prediction lengths $O \in \{96, 192, 336, 720\}$. Performance is evaluated based on the number of parameters (in millions) and MSE.}
    \label{tbl:model-complexity}
    \begin{tabular*}{\tblwidth}{@{}CC|CC|CC@{}}
        \toprule
        \multicolumn{2}{l}{Methods} & \multicolumn{2}{c}{CANet w/ SKPL} & \multicolumn{2}{c}{CANet w/o SKPL} \\
        \cmidrule(lr) {3-4} \cmidrule(lr) {5-6}
        \multicolumn{2}{l}{Metric} & \multicolumn{1}{c}{\# Params} & \multicolumn{1}{c}{MSE} & \multicolumn{1}{c}{\# Params} & \multicolumn{1}{c}{MSE} \\
        \midrule
        \multirow{5}{*}{\begin{turn}{90}\textit{ETTm2}\end{turn}} & 96 & 0.60 & 0.173 & 1.02 & 0.172 \\
        & 192 & 1.09 & 0.236 & 1.91 & 0.238 \\
        & 336 & 0.84 & 0.287 & 1.27 & 0.288 \\
        & 720 & 1.85 & 0.371 & 2.70 & 0.370 \\
        \midrule
        \multirow{5}{*}{\begin{turn}{90}\textit{Exchange}\end{turn}} & 96 & 0.53 & 0.078 & 0.94 & 0.078 \\
        & 192 & 2.09 & 0.157 & 3.73 & 0.159 \\
        & 336 & 2.30 & 0.276 & 4.00 & 0.281 \\
        & 720 & 1.66 & 0.490 & 2.51 & 0.490 \\
        \midrule
        \multicolumn{2}{l}{Avg.} & \multicolumn{1}{c}{1.37} & \multicolumn{1}{c}{0.259} & \multicolumn{1}{c}{2.26} & \multicolumn{1}{c}{0.260} \\
        \bottomrule
    \end{tabular*}
\end{table}

\subsection{Model complexity analysis}
\label{sec:model-complexity}

We carried out a comprehensive comparison of CANet and baseline models from Section \ref{subsec:baselines} in terms of forecasting accuracy, single-input inference speed, and parameter count, as shown in Figure \ref{fig:model-complexity}. In order to ensure consistency, we used each model’s official configurations and performed inference tests on single entries, averaging the results. Model efficiency was evaluated on two representative datasets, ETTm2 and Exchange, as shown in Figure \ref{fig:model-complexity}. Tests were conducted using look-back windows of 96, 192, 336, and 720 time steps, with results averaged across these intervals.

Overall, CANet achieves competitive performance across these metrics. Notably, in addition to delivering top-tier forecasting accuracy, CANet’s parameter count and inference speed fall well below the average, with approximately \textbf{70\%} fewer parameters and an \textbf{80\%} faster inference time compared to the baseline models. These findings highlight CANet's lightweight architecture, emphasizing its suitability for efficient, deployable time series analysis.

To further validate the impact of the Stacked Kronecker Product Layer (SKPL) within CANet, we conducted an additional analysis comparing the model’s complexity and performance with and without SKPL, as presented in Table \ref{tbl:model-complexity}. In this comparison, SKPL was replaced with a traditional linear layer. For a fair comparison, we used the same hyperparameters for the CANet versions without SKPL with the CANet models from Table \ref{tbl:forecasting-results}, except for the learning rate and dropout rate, which are tuned separately. For the CANet versions without SKPL, we performed a search for the optimal learning rate and dropout rate using Optuna, conducting 50 training runs since we focused solely on these two parameters. According to our findings, incorporating the Stacked Kronecker Product Layer reduces the parameter count by an average of \textbf{39.3\%} (from 2.26 million to 1.37 million) without any loss in performance.

\begin{table*}[width=1.75\linewidth,cols=12]
    \caption{CANet's performance across varying noise levels $\alpha$ with a look-back window $L$ of 96 and prediction lengths $O \in \{96, 192, 336, 720\}$.}
    \label{tbl:noise-robustness}
    \begin{tabular*}{\tblwidth}{@{}CC|CC|CC|CC|CC|CC@{}}
        \toprule
        \multicolumn{2}{l}{Noise Level} & \multicolumn{2}{c}{$\alpha = 0.1$} & \multicolumn{2}{c}{$\alpha = 0.2$} & \multicolumn{2}{c}{$\alpha = 0.3$} & \multicolumn{2}{c}{$\alpha = 0.4$} & \multicolumn{2}{c}{$\alpha = 0.5$} \\
        \cmidrule(lr) {3-4} \cmidrule(lr) {5-6} \cmidrule(lr) {7-8} \cmidrule(lr) {9-10} \cmidrule(lr) {11-12}
        \multicolumn{2}{l}{Metric} & \multicolumn{1}{c}{MSE} & \multicolumn{1}{c}{MAE} & \multicolumn{1}{c}{MSE} & \multicolumn{1}{c}{MAE} & \multicolumn{1}{c}{MSE} & \multicolumn{1}{c}{MAE} & \multicolumn{1}{c}{MSE} & \multicolumn{1}{c}{MAE} & \multicolumn{1}{c}{MSE} & \multicolumn{1}{c}{MAE}\\
        \midrule
        \multirow{5}{*}{\begin{turn}{90}\textit{ETTh2}\end{turn}} & 96 & 0.230 & 0.306 & 0.232 & 0.310 & 0.236 & 0.317 & 0.241 & 0.323 & 0.246 & 0.332 \\
        & 192 & 0.291 & 0.345 & 0.292 & 0.348 & 0.295 & 0.354 & 0.299 & 0.360 & 0.304 & 0.367 \\
        & 336 & 0.334 & 0.377 & 0.336 & 0.380 & 0.338 & 0.385 & 0.342 & 0.39 & 0.346 & 0.397 \\
        & 720 & 0.397 & 0.426 & 0.397 & 0.429 & 0.400 & 0.434 & 0.406 & 0.442 & 0.414 & 0.452 \\
        \cmidrule(lr) {2-12}
        & Avg. & 0.313 & 0.364 & 0.314 & 0.367 & 0.317 & 0.373 & 0.322 & 0.379 & 0.328 & 0.387 \\
        \midrule
        \multirow{5}{*}{\begin{turn}{90}\textit{Weather}\end{turn}} & 96 & 0.178 & 0.242 & 0.190 & 0.259 & 0.194 & 0.276 & 0.206 & 0.293 & 0.219 & 0.310 \\
        & 192 & 0.216 & 0.27 & 0.223 & 0.282 & 0.227 & 0.290 & 0.231 & 0.296 & 0.234 & 0.302 \\
        & 336 & 0.256 & 0.313 & 0.264 & 0.328 & 0.271 & 0.340 & 0.279 & 0.352 & 0.288 & 0.362 \\
        & 720 & 0.340 & 0.373 & 0.350 & 0.394 & 0.364 & 0.416 & 0.386 & 0.437 & 0.410 & 0.457 \\ 
        \cmidrule(lr) {2-12}
        & Avg. & 0.248 & 0.300 & 0.256 & 0.316 & 0.264 & 0.331 & 0.276 & 0.345 & 0.288 & 0.358 \\
        \midrule
        \bottomrule
    \end{tabular*}
\end{table*}

\begin{figure*}[]
    \centering
    \includegraphics[width=0.9\linewidth]{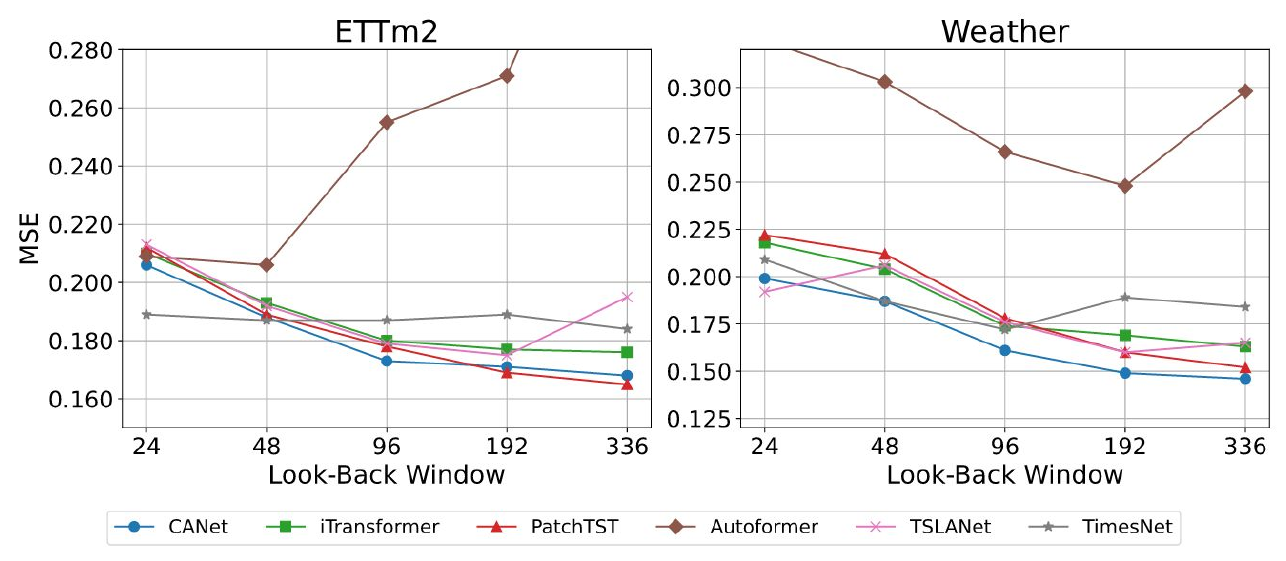}
    \caption[]{Model performance comparison across varying look-back windows $L \in \{24, 48, 96, 192, 336\}$ on the ETTm2 and Weather datasets, with a fixed prediction length $O = 96$ for each model.}
    \vspace{-5mm}
    \label{fig:varying-lookback-window}
\end{figure*}

\subsection{Model component analysis}
\label{sec:model-component}

To better understand the source of CANet's performance, we conducted an ablation study by individually removing its core components: Multi-Resolution Patching (MRP), Adaptive Spectral Block (ASB), Interactive Convolutional Block (ICB), and the Style Blending Gate (BG) within the NSAN module. Experiments were performed on the ETTm2 and Exchange datasets, and the results are summarized in Table \ref{tbl:ablation}. We presented the results after tuning the hyperparameters as in other experiments.

The removal of the Style Blending Gate resulted in the largest performance drop, increasing MSE from 0.267 to 0.277 on ETTm2, demonstrating that CANet’s ability to reintegrate non-stationary dynamics depends on this gate, which adaptively fuses internal and external statistics.. Replacing MRP with a fixed patch size, 16, also degraded performance, increasing MSE from 0.250 to 0.258 on Exchange. This confirms the value of multi-resolution patching. Disabling the ASB led to a moderate increase in error, increasing MSE from 0.250 to 0.256 on Exchange, indicating its effectiveness in denoising and frequency-based filtering. Lastly, removing ICB caused a smaller but consistent decline, suggesting its contribution to capturing local interactions.

These results demonstrate that CANet’s performance stems from the synergy among its components rather than any single module. The combination of these components plays a central role in achieving robust forecasting.

\subsection{Varying look-back window}
\label{subsec:varying-lookback-window}

Previous studies demonstrate that forecasting accuracy does not always increase with longer look-back windows \citep{Nie2023, Zeng2023}, which contrasts with theoretical expectations based on the statistical methods \citep{Box1968}. In theory, models with extended look-back windows should improve performance by utilizing more historical data to predict future values. In order to assess this in practice, we compared CANet’s forecasting accuracy with some of the baseline models across varying look-back window lengths.

In particular, we evaluated the forecasting performance on the ETTm2 and Weather datasets, testing look-back window length $L \in \{24, 48, 96, 192, 336\}$. The results, illustrated in Figure \ref{fig:varying-lookback-window} and summarized in Table \ref{tbl:varying-lookback-window}, reveal that CANet’s performance smoothly improves with longer look-back windows. CANet achieves its highest accuracy with extended windows, showcasing its capability to effectively leverage extended historical data for forecasting. This result highlights CANet’s strength in capturing relevant temporal patterns across varying historical contexts.

As in Section \ref{sec:model-complexity}, we conducted a focused search for the optimal learning rate and dropout rate using Optuna, with 50 training runs to tune these parameters.

\begin{table*}
\footnotesize
\centering
\caption{CANet forecasting accuracy on \textit{ETTh1/2}, \textit{ETTm1/2}, and \textit{Electricity}. Values are mean $\pm$ standard deviation over five independent runs. All datasets share the four canonical prediction horizons $O\!\in\!\{96,192,336,720\}$.}
\label{tbl:stats1}
\renewcommand{\arraystretch}{1.25}
{\setlength{\tabcolsep}{3pt}
\scalebox{0.9}{
\begin{tabular}{cc|cc|cc|cc|cc|cc}
\toprule
\multicolumn{2}{c}{Dataset} &
\multicolumn{2}{c}{\textit{ETTh1}} &
\multicolumn{2}{c}{\textit{ETTh2}} &
\multicolumn{2}{c}{\textit{ETTm1}} &
\multicolumn{2}{c}{\textit{ETTm2}} &
\multicolumn{2}{c}{\textit{Electricity}} \\ 
\cmidrule(lr) {3-4} \cmidrule(lr) {5-6} \cmidrule(lr) {7-8} \cmidrule(lr) {9-10} \cmidrule(lr) {11-12}
\multicolumn{2}{c}{Metric} &
MSE & MAE & MSE & MAE & MSE & MAE & MSE & MAE &
MSE & MAE \\
\midrule
\multirow{5}{*}{}
& 96 & 0.359$\pm$0.004 & 0.389$\pm$0.001 & 0.231$\pm$0.002 & 0.304$\pm$0.001 &
             0.321$\pm$0.003 & 0.361$\pm$0.002 & 0.183$\pm$0.006 & 0.265$\pm$0.005 &
             0.163$\pm$0.000 & 0.253$\pm$0.001 \\
& 192 & 0.390$\pm$0.008 & 0.407$\pm$0.002 & 0.294$\pm$0.004 & 0.346$\pm$0.002 &
             0.363$\pm$0.005 & 0.385$\pm$0.003 & 0.242$\pm$0.004 & 0.303$\pm$0.004 &
             0.166$\pm$0.000 & 0.263$\pm$0.001 \\
& 336 & 0.429$\pm$0.018 & 0.435$\pm$0.009 & 0.377$\pm$0.031 & 0.396$\pm$0.013 &
             0.378$\pm$0.005 & 0.405$\pm$0.003 & 0.292$\pm$0.003 & 0.336$\pm$0.002 &
             0.187$\pm$0.002 & 0.281$\pm$0.002 \\
& 720 & 0.470$\pm$0.012 & 0.467$\pm$0.002 & 0.413$\pm$0.010 & 0.436$\pm$0.004 &
             0.436$\pm$0.005 & 0.434$\pm$0.003 & 0.374$\pm$0.002 & 0.386$\pm$0.001 &
             0.229$\pm$0.004 & 0.316$\pm$0.002 \\
\cline{2-12}
& Avg      & 0.412$\pm$0.011 & 0.425$\pm$0.004 & 0.329$\pm$0.012 & 0.371$\pm$0.005 &
             0.375$\pm$0.005 & 0.396$\pm$0.003 & 0.273$\pm$0.004 & 0.323$\pm$0.003 &
             0.186$\pm$0.002 & 0.278$\pm$0.002 \\
\bottomrule
\end{tabular}}}
\end{table*}

\begin{table*}
\footnotesize
\centering
\caption{CANet forecasting accuracy on \textit{Exchange}, \textit{Weather}, \textit{ILI}, and \textit{Traffic}. Metrics are reported as mean $\pm$ standard deviation over five seeds.  
Horizons are $O\!\in\!\{96,192,336,720\}$ for every dataset except \textit{ILI}, whose horizons are $\{24,36,48,60\}$.}
\label{tbl:stats2}
\renewcommand{\arraystretch}{1.25}
{\setlength{\tabcolsep}{3pt}
\scalebox{0.9}{
\begin{tabular}{cc|cc|cc|cc|cc}
\toprule
\multicolumn{2}{c}{Dataset} &
\multicolumn{2}{c}{\textit{Exchange}} &
\multicolumn{2}{c}{\textit{Weather}} &
\multicolumn{2}{c}{\textit{ILI}} &
\multicolumn{2}{c}{\textit{Traffic}} \\ 
\cmidrule(lr) {3-4} \cmidrule(lr) {5-6} \cmidrule(lr) {7-8} \cmidrule(lr) {9-10}
\multicolumn{2}{c}{Metric} &
MSE & MAE & MSE & MAE & MSE & MAE & MSE & MAE \\
\midrule
\multirow{5}{*}{}
& 96 (24)  & 0.079$\pm$0.001 & 0.198$\pm$0.002 & 0.161$\pm$0.001 & 0.210$\pm$0.001 &
             1.576$\pm$0.073 & 0.791$\pm$0.021 & 0.534$\pm$0.008 & 0.339$\pm$0.003 \\
& 192 (36) & 0.177$\pm$0.021 & 0.306$\pm$0.019 & 0.205$\pm$0.001 & 0.251$\pm$0.001 &
             1.774$\pm$0.272 & 0.824$\pm$0.067 & 0.518$\pm$0.009 & 0.329$\pm$0.004 \\
& 336 (48) & 0.304$\pm$0.028 & 0.406$\pm$0.015 & 0.246$\pm$0.001 & 0.288$\pm$0.002 &
             2.109$\pm$0.569 & 0.920$\pm$0.123 & 0.540$\pm$0.007 & 0.341$\pm$0.002 \\
& 720 (60) & 0.561$\pm$0.049 & 0.571$\pm$0.022 & 0.327$\pm$0.003 & 0.343$\pm$0.003 &
             2.013$\pm$0.168 & 0.926$\pm$0.043 & 0.570$\pm$0.018 & 0.353$\pm$0.001 \\
\cline{2-10}
& Avg      & 0.280$\pm$0.011 & 0.370$\pm$0.015 & 0.235$\pm$0.002 & 0.273$\pm$0.002 &
             1.868$\pm$0.271 & 0.865$\pm$0.064 & 0.541$\pm$0.011 & 0.341$\pm$0.003 \\
\bottomrule
\end{tabular}}}
\end{table*}

\begin{table}[width=0.9\linewidth,cols=4]
    \caption{Results of CANet for different look-back window lengths. Prediction length $O$ is fixed at 96, with look-back window $L \in \{24, 48, 96, 192, 336\}$.}
    \label{tbl:varying-lookback-window}
    \begin{tabular*}{\tblwidth}{@{}CC|CC@{}}
        \toprule
        \multicolumn{2}{l}{Dataset} & \multicolumn{1}{c}{MSE} & \multicolumn{1}{c}{MAE} \\
        \midrule
        \multirow{5}{*}{\textit{ETTm2}} & 24 & 0.206 & 0.286 \\
           & 48 & 0.188 & 0.272 \\
           & 96 & 0.173 & 0.258 \\
           & 192 & 0.171 & 0.255 \\
           & 336 & 0.168 & 0.253 \\
        \midrule
        \multirow{5}{*}{\textit{Weather}} & 24 & 0.199 & 0.243 \\
           & 48 & 0.187 & 0.235 \\
           & 96 & 0.161 & 0.211 \\
           & 192 & 0.149 & 0.196 \\
           & 336 & 0.146 & 0.198 \\
        \midrule
        \bottomrule
    \end{tabular*}
\end{table}

\subsection{Robustness against noisy data}
\label{subsec:robustness-against-noisy-data}

To validate CANet's performance in adaptive noise-filtering provided by the Adaptive Spectral Block \citep{Eldele2024}, we tested its forecasting accuracy on data with varying levels of added noise. We generated progressively noisier versions of the original input data by introducing Gaussian noise at levels of 0.1, 0.2, and 0.4, formalized as follows:

\begin{equation}
    X'_{\textit{input}} = X_{\textit{input}} + \alpha \times \mu
\end{equation}

where $X_{\textit{input}}$ represents the original input, $\alpha$ indicates the noise level and $\mu$ is a random variable drawn from a Gaussian distribution with a mean of 0 and variance of 1.

The results, shown in Table \ref{tbl:noise-robustness}, reveal that CANet consistently demonstrates robustness against noise across various benchmarks (ETTh2 and Weather). Although its forecasting accuracy declines as noise levels increase, the degradation remains within acceptable limits. Particularly in the presence of lower-scale noise (e.g., from 0.1 to 0.5), CANet retains its predictive accuracy and stability, underscoring its reliability for multivariate long-term time series forecasting even under noisy conditions.

\begin{table}
\footnotesize
\centering
\caption{Paired t-test results comparing CANet and TSLANet over five seeds and four horizons per dataset (20 paired evaluations). 
$\Delta$MSE / $\Delta$MAE denote CANet $-$ TSLANet (negative values indicate lower error for CANet). 
\textbf{Bold} entries indicate statistically significant improvements at one-sided $p<0.05$.}
\label{tbl:paired_ttest}
\renewcommand{\arraystretch}{1.25}
{\setlength{\tabcolsep}{6pt}
\scalebox{0.9}{
\begin{tabular}{l|cccc}
\toprule
Dataset & $\Delta$MSE & $p_{\text{MSE}}$ & $\Delta$MAE & $p_{\text{MAE}}$ \\
\midrule
ETTh1        & -0.034 & $\boldsymbol{8.40\times10^{-9}}$  & -0.015 & $\boldsymbol{5.32\times10^{-7}}$  \\
ETTh2        & -0.026 & $\boldsymbol{6.16\times10^{-3}}$  & -0.020 & $\boldsymbol{7.17\times10^{-4}}$  \\
ETTm1        & -0.016 & $\boldsymbol{2.73\times10^{-7}}$  & -0.005 & $\boldsymbol{1.19\times10^{-6}}$  \\
ETTm2        & -0.011 & $\boldsymbol{1.05\times10^{-3}}$  & -0.005 & $\boldsymbol{8.19\times10^{-3}}$  \\
Electricity  & $-0.013$ & $\boldsymbol{1.13\times10^{-14}}$ & -0.004 & $\boldsymbol{9.57\times10^{-7}}$  \\
Exchange     & -0.080 & $\boldsymbol{5.96\times10^{-3}}$  & -0.033 & $\boldsymbol{9.63\times10^{-3}}$  \\
ILI          & -0.129          & $1.09\times10^{-1}$      & -0.024 & $1.42\times10^{-1}$           \\
Traffic      & 0.048           & $1.00$      & 0.026 & $1.00$           \\
Weather      & -0.024 & $\boldsymbol{4.44\times10^{-12}}$ & \textbf{$-0.006$} & $\boldsymbol{2.92\times10^{-10}}$ \\
\bottomrule
\end{tabular}}}
\end{table}

\subsection{Cross-architecture comparison}
\label{subsec:cross-arch}

To assess whether CANet's performance could be replicated by simply adding normalization schemes to an existing architecture, we conducted a series of experiments with TSLANet. Specifically, we tested three variants: TSLANet with RevIN, AdaIN, and NSAN module. These models were evaluated under the same experimental settings suggested for TSLANet, and the results are summarized in Table \ref{tbl:ablation}.

The findings reveal that none of the modified TSLANet variants matched CANet's performance. TSLANet with RevIN and AdaIN consistently underperformed, particularly on the highly non-stationary Exchange dataset. While TSLANet with NSAN achieved better results than the other two variants, it still lagged behind CANet by a margin of approximately 6\% to 30\% in MSE, depending on the dataset and horizon.

These results suggest that CANet's improvements cannot be attributed to normalization alone. Instead, its strength lies in the tight integration of NSAN with CANet's architectural components. Furthermore, the Style Blending Gate in NSAN, which adaptively fuses internal and external statistics, plays a key role in conditioning feature representations on non-stationary temporal patterns, a behavior that simpler normalization methods like RevIN or standalone AdaIN cannot replicate.

\subsection{Statistical robustness}
\label{subsec:stats}

To verify CANet's performance in various independent initialisations, we re-executed every experiment with five different random seeds (56, 42, 0, 2, and 2025) with the hyperparameter setup used in Section \ref{subsec:long-term-multivariate-time-series-forecasting}. Table \ref{tbl:stats1} and Table \ref{tbl:stats2} report the mean and standard deviation ($\mu\pm\sigma$) of MSE and MAE over these runs for all prediction horizons.

Across the 72 dataset–horizon pairs, the median standard deviation is $\sigma_{MSE}=0.005$ and $\sigma_{MAE}=0.003$. Such tight error bands confirm that CANet's gains are considerably stable. Additionally, to rigorously quantify whether CANet's improvements are statistically reliable despite high variance observed in some datasets, we performed paired t-tests against TSLANet. TSLANet was selected as the comparative baseline because it serves as the architectural backbone of CANet; thus, this paired testing isolates the specific contributions--especially NSAN-- while controlling for seed-dependent initialization noise.

The paired analysis (Table \ref{tbl:paired_ttest}) shows that CANet obtains statistically significant improvements ($p<0.05$) on seven datasets: ETTh1, ETTh2, ETTm1, ETTm2, Electricity, Exchange, and Weather. On ILI, the differences are not statistically significant, and on Traffic CANet performs worse; these cases are reported transparently.

Overall, the expanded multi-seed evaluation and paired significance tests demonstrate that CANet's improvements are consistent and statistically robust.

\section{Conclusion}
\label{sec:conclusion}

This paper introduces CANet, a novel framework designed for multivariate long-term time series forecasting that addresses the limitations posed by the non-stationary nature of data. Different from previous approaches that either reduce non-stationarity at the risk of over-stationarization or lack deployability due to complexity, CANet offers a lightweight and practical solution. At its core, CANet introduces an innovative Non-stationary Adaptive Normalization (NSAN) module, which reintegrates non-stationary factors back into the model. This mechanism not only enhances data predictability but also significantly improves the model's forecasting accuracy by capturing and leveraging these dynamic factors within time series data. 

Our experimental results demonstrate that CANet consistently outperforms other baseline models across several widely-used multivariate time series forecasting benchmarks. Detailed ablation studies further validate the effectiveness of the components within our proposed model. The key contributions of this research lie in developing a more accurate, efficient model architecture that avoids the pitfalls of over-stationarization while remaining lightweight and easy to deploy.

Looking ahead, future work will focus on refining CANet's internal structure to make it more versatile, potentially enabling its integration with a broader range of baseline models in the time series forecasting domain.

\section*{Acknowledgements}
This research received funding from the Research Universities Support Program (YOK-ADEP) with project number ADEP-312-2024-11490.

\printcredits

\section*{Declaration of competing interest}
The authors declare that they have no known competing financial interests or personal relationships that could have appeared to
influence the work reported in this paper.

\section*{Data availability}
Data will be made available on request.

\bibliographystyle{cas-model2-names}

\bibliography{cas-refs}



\end{document}